\documentclass{article}

    \PassOptionsToPackage{numbers, compress}{natbib}
\usepackage{multirow} 
\usepackage{algorithm}
\usepackage{algpseudocode} 
\usepackage{tikz}
\usepackage{amsmath}
\usepackage{amsthm,amssymb}
\usepackage{tabularx,array,booktabs}
\usepackage{graphicx}
\usepackage{thm-restate} 
\makeatletter\newcommand{\firstonly}[1]{\ifthmt@thisistheone#1\fi}\makeatother 
\newtheorem{theorem}{Theorem}[section]

\theoremstyle{plain}

\theoremstyle{definition}
\newtheorem{definition}[theorem]{Definition}

\theoremstyle{remark}
\newtheorem{remark}[theorem]{Remark}

 \usepackage[main, final]{neurips_2026}

\usepackage[utf8]{inputenc} 
\usepackage[T1]{fontenc}    
\usepackage{hyperref}       
\usepackage{url}            
\usepackage{booktabs}       
\usepackage{amsfonts}       
\usepackage{nicefrac}       
\usepackage{microtype}      
\usepackage{xcolor}         
\usepackage{enumitem}
\usepackage{blindtext}

\newcommand{\qh}[1]{\textcolor{black}{#1}}
\newcommand{\camera}[1]{{\color{black!50!black}#1}}
\newcommand{\cameracolor}{\color{black!50!black}} 

\title{Optimal Subgroup Discovery at \\ Every Support Threshold}

\author{%
  Lincen Yang$^{1}$\thanks{Equal contribution.}
  \And
  Qi Huang$^{1}$\footnotemark[1]
  \And
  Niki van Stein$^{1}$
  \AND
  Thomas B\"{a}ck$^{1}$
  \And
  Matthijs van Leeuwen$^{1}$
  \And
  Zhong Li$^{2,1}$\thanks{Corresponding author.}
  \AND
  \mdseries
  $^{1}$LIACS, Leiden University, The Netherlands \\
  $^{2}$Great Bay University, China \\
  \texttt{\{l.yang, q.huang, n.van.stein, t.h.w.baeck, } \\
  \texttt{m.van.leeuwen, z.li\}@liacs.leidenuniv.nl}
}

\begin{document}

\maketitle

\begin{abstract}
Subgroup discovery aims to identify regions of the feature space where a target variable deviates from its marginal distribution. A fundamental challenge in this problem is the trade-off between a subgroup's \emph{deviation} and its \emph{support}: small subgroups can be highly deviating but statistically meaningless, while large ones cannot deviate much by construction. Existing methods tackle this challange by collapsing the two into a single scalar quality measure, implicitly committing to one particular trade-off and making it difficult to recover the most deviating subgroup under a user-specified support constraint. We instead provide the first theoretical and structural characterization of optimal subgroups with maximum Kullback-Leibler (KL) divergence under arbitrary support constraints. Under a partition-based generative model, where the feature space decomposes into regions of homogeneous conditional distributions of the target variable called \emph{atoms}, we prove that every point on the deviation-support Pareto front is a union of whole atoms together with at most one fractional atom. This reduces an uncountable search to a finite combinatorial problem. Guided by this theory, we propose a simple two-step recipe: learn a homogeneous partition, then select atoms greedily. Across 19 benchmark datasets, it finds the most deviating subgroups \camera{at} a range of support thresholds, and reliably improves subgroups returned by existing methods when applied as a post-processing step. These results empirically validate the sufficiency and necessity of our theoretical contributions.
\end{abstract}

\section{Introduction} \label{sec:intro}
Subgroup discovery is the task of finding subsets of the feature space such that some target variable deviates from its marginal distribution. It has been successfully applied across various domains, and recent examples include knowledge discovery in heterogeneous catalysts in chemistry~\citep{esterhuizen2022interpretable} and recommendation systems~\citep{iferroudjene2023methods}. As the goal is often to obtain insights from data, subgroups are typically described by rules or unions of rules, where each rule is a  conjunction of literals~\citep{furnkranz2012foundations}. For example, in a dataset of housing prices with features such as location, size, and age, a subgroup might be ``(location $=$ coastal) $\wedge$ (size $\geq 200 \text{m}^2$)'', which isolates a region of the feature space in which the conditional distribution of price differs markedly from the overall distribution. 

Formally, given a feature vector $X \in \mathcal{X}$ and target $Y \in \mathcal{Y}$ distributed according to some joint distribution $P$, a subgroup is any measurable subset $Q \subseteq \mathcal{X}$, and its quality is measured by how much the conditional distribution $P(Y \mid X \in Q)$ deviates from the marginal $P(Y)$, typically via a divergence such as the Kullback-Leibler (KL) divergence, and by how large the subgroup is, measured by its support $P(X\in Q)$. 

\textbf{The support-deviation trade-off.}
A fundamental challenge in subgroup discovery is that deviation and support pull in opposite directions: very small subgroups can be arbitrarily deviating but statistically meaningless, while large ones cannot deviate much by construction. Practitioners need subgroups that are both deviating enough to be informative and large enough to be actionable.

\textbf{Existing work conflates this trade-off into a single score.}
The dominant approach to handling this challenge is to fold support and deviation into a single scalar quality measure. Classic measures include weighted relative accuracy, weighted mean differences, and support-weighted KL divergence~\citep{lavravc2004subgroup, klosgen1996explora, van2010maximal}. Recent state-of-the-art methods retain this design: RSD~\citep{proencca2022robust} penalizes small subgroups via an MDL description-length term, and SyFlow~\citep{xu2024learning} maximizes a support-weighted KL objective through differentiable neural optimization. While this single-score strategy enables unified optimization, it implicitly commits to one particular support--deviation trade-off, and different weighting choices yield different subgroups.

\begin{figure}[ht]
    \centering
\includegraphics[width=0.9\linewidth]{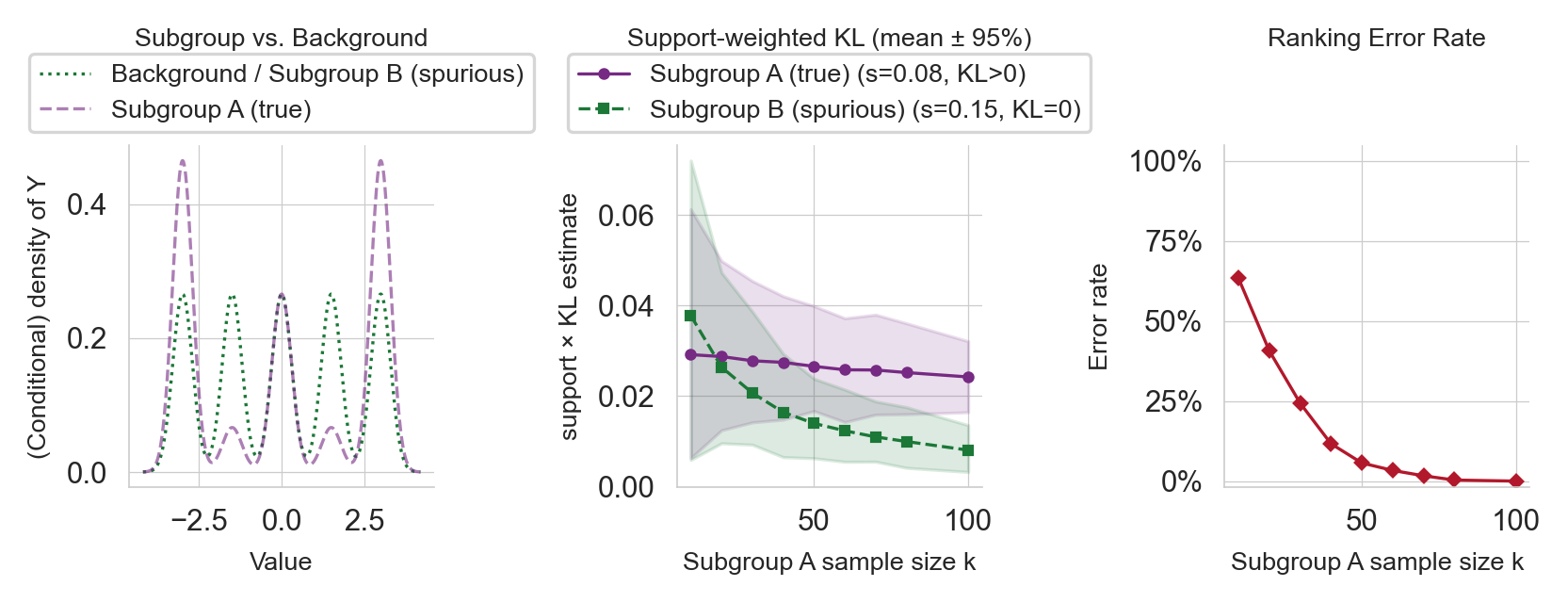}
    \caption{Ranking error under support-weighted KL estimation with finite samples. Two subgroups are drawn from a 5-mode Gaussian mixture background: Subgroup~A (true, support $= 0.08$, $\mathrm{KL} \approx 0.25$) and Subgroup~B (spurious, support $= 0.15$, $\mathrm{KL} = 0$). \textbf{Left:} Conditional densities of the subgroups and background. \textbf{Middle:} Support-weighted KL estimates (support $\times$ KL) as a function of Subgroup~A's sample size~$k$, with B's size scaled by the support ratio; shaded regions show the 2.5th--97.5th percentile range over 300 draws. At small~$k$, estimation noise inflates B's score above A's despite its zero true divergence. \textbf{Right:} Fraction of trials in which the spurious Subgroup~B is ranked above the true Subgroup~A. At $k = 30$ the error rate still exceeds 20\%, showing that support-weighted KL ranking is unreliable at small sample sizes.}
    \label{fig:rankflip}
\end{figure}

More fundamentally, support-weighting does not actually resolve the underlying problem. Weighted scores such as the weighted-KL-divergence discourage small subgroups by penalizing coverage, on the implicit assumption that larger subgroups yield more reliable KL estimates. However, multiplying an unreliable estimate by its support merely rescales the noise, and because the weight asymmetrically rewards high-support subgroups, upward noise in a high-support subgroup's KL estimate can translate into a ranking advantage over genuinely deviating low-support ones. Figure~\ref{fig:rankflip} illustrates this: with two subgroups drawn from a Gaussian-mixture background---one (Subgroup~A) genuinely deviating at support~$0.08$, the other (Subgroup~B) a pure sample from the background at support~$0.15$---the support-weighted KL ranks the spurious Subgroup~B above the genuine Subgroup~A more than 20\% of the time at sample size $k=30$. Full reproducibility details are in Appendix~\ref{appendix:sec:exp_intro}.


\textbf{Research gap.}
What is missing is a rigorous understanding of what makes a subgroup maximally deviating \emph{given an arbitrary support constraint}, together with a practical recipe for identifying the entire deviation-support Pareto front~\citep{emmerich2018tutorial}. That is, rather than committing to a single implicit trade-off via a weighted score, one would ideally like to ask: \emph{for any user-specified minimum support $\alpha$, what is the most deviating subgroup?} Answering this question in its full generality has remained open: existing exhaustive algorithms with guarantees~\citep{grosskreutz2008tight, lemmerich2013difference} and recent gradient-based methods~\citep{xu2024learning} all target weighted objectives rather than the constrained problem directly.

\textbf{Our approach.}
We approach subgroup discovery from a probabilistic modeling perspective. Our key insight is that when the data-generating distribution admits a partition-based structure where the feature space decomposes into regions with homogeneous conditional distributions $P(Y|X)$, which we refer to as \emph{atoms}, the search for optimal subgroups at \emph{every} support level collapses to a combinatorial search problem over a finite set of atoms. Specifically, we prove that 
under any support constraint, the optimal subgroup is a union of whole atoms together with at most one fractional atom. This structural result characterizes the full deviation-support Pareto front as a discrete selection problem over atoms that are determined by the distribution alone, independent of the support threshold. This suggests a clean two-step recipe: first learn a partition of the feature space into atoms with (approximately) homogeneous conditional distributions of the target, then select atoms to optimize deviation under any desired support constraint.

\textbf{Contributions.} Our contributions can be summarized as follows.
\begin{itemize}
    \item We develop, to our knowledge, the first theoretical and structural characterization of KL-optimal subgroups under support constraints. Under a partition-based generative model, we prove that every point on the deviation-support Pareto front is a union of whole atoms plus at most one fractional atom (Theorem~\ref{thm:support_existence}). Under distinct generating distributions, this structure is also necessary (Corollary~\ref{cor:uniqueness}).
    \item Our theory suggests a modular recipe for learning optimal subgroups from data, i.e., estimate a homogeneous partition, then select atoms, which reduces an uncountable search space to a finite combinatorial problem that can be solved at any user-specified support threshold.
    \item We instantiate the recipe with a partition estimator (CDTree~\citep{yang2024conditional}) and a greedy atom-selection algorithm (GAP), which together form the \emph{AtomSD} framework. Across 19 benchmark datasets, AtomSD achieves the best average rank across a wide range of support thresholds, empirically validating the \emph{sufficiency} of our theoretical results. When applied as a post-processing step, AtomSD also reliably improves subgroups returned by four existing methods, validating the \emph{necessity} condition implied by our theorem.
\end{itemize}

\section{Related work}

Subgroup discovery was introduced by~\citet{klosgen1996explora} and~\citet{wrobel1997algorithm}, building on an earlier formalization under the name data surveying~\citep{siebes1995data}. Early work targeted discrete targets, where probability distributions are easier to estimate~\citep{lavravc2004subgroup, kavvsek2006apriori, abudawood2009evaluation}. Common quality measures include weighted accuracy and its variants~\citep{lavravc2004subgroup, kavvsek2006apriori}, weighted relative accuracy~\citep{abudawood2009evaluation}, KL divergence for discrete variables~\citep{van2010maximal}, and weighted mean differences~\citep{klosgen1996explora}; we refer to~\citet{atzmueller2015subgroup} and~\citet{herrera2011overview} for comprehensive surveys. Further, a separate strand evaluates entire subgroup sets rather than individual subgroups, incorporating diversity between subgroups~\citep{matthijs2012diverse, bosc2018anytime, belfodil2019fssd}.

Recent work has shifted to the harder case of numeric targets. Some methods focus on specific quality measures such as weighted KL divergence and its variants~\citep{proencca2020discovering, proencca2022robust, xu2024learning, boley2017identifying, kalofolias2017efficiently}; others propose frameworks compatible with a range of measures~\citep{lemmerich2016fast}. All such measures fold support and deviation into a single score, committing implicitly to one trade-off. Our work instead targets the support-constrained problem directly: for any user-specified support threshold, find the most deviating subgroup. This yields the full support--deviation Pareto front~\citep{emmerich2018tutorial} rather than a single point on it.

Most existing methods assume one-dimensional targets, but exceptional model mining methods handle multi-dimensional ones~\citep{leman2008exceptional, duivesteijn2010subgroup, duivesteijn2016exceptional}. Our theorems do not require one-dimensional targets and apply in principle to this setting, though our empirical validation focuses on one-dimensional cases.

A parallel research line focuses on algorithms. Most are heuristic discrete-optimization procedures~\citep{proencca2022robust, matthijs2012diverse, lavravc2004subgroup, meeng2021real, proencca2020discovering}. More recent work represents subgroup discovery as a neural network optimized by stochastic gradient~\citep{xu2024learning}, approximating the discrete problem with a continuous one. Algorithms with guarantees also exist, based on optimistic estimates~\citep{wrobel1997algorithm, lemmerich2016fast, lemmerich2013difference, grosskreutz2008tight} or FP-growth~\citep{atzmueller2006sd, atzmueller2009fast, boley2017identifying, kalofolias2017efficiently}. Our work is complementary to them as these algorithms all optimize quality measures that fold support and deviation together, so they do not directly solve the support-constrained problem we target.

\section{Theory} \label{sec:theory}
We consider a supervised setting where the feature $X$ can be continuous, discrete, or a mix of both, and the target $Y$ can be binary, multi-class, or continuous. A subgroup is any measurable subset $Q \subseteq \mathcal{X}$, and we are interested in those whose conditional target distribution deviates from the population. We use $P_X$, $P_Y$, and $P_{Y|X}$ to denote the marginal distributions of $X$, $Y$, and the conditional distribution of $Y$ given $X$, respectively. \camera{Throughout the development of our theory,} we work with population-level probability distributions. 

The degree of deviation of a subgroup $Q$ is measured by the KL-divergence:
\begin{equation} \label{eq:deviation}
    d(Q) = \mathrm{KL}\!\left(P(Y \mid X\in Q) \,\|\, P(Y)\right), \quad \forall\, Q \subseteq \mathcal{X},\; P(X\in Q)>0.
\end{equation}
The goal of subgroup discovery is to find the subgroup with the largest \camera{deviation}, possibly with the support constraint that $P(X\in Q) \geq \alpha$. That is, in practice, analysts and users often look for subgroups large enough to be actionable (e.g., a subgroup of hospital patients large enough to support a future clinical trial). For interpretability, subgroup discovery methods commonly express $Q$ as a rule or union of rules, where a \emph{rule} is a conjunction of the form $\bigwedge_{j \in J}(X_j \in R_j)$ with each $R_j$ an interval (if $X_j$ is continuous) or a discrete set. Our theoretical results do not depend on this restriction, though it motivates Definition~\ref{def:partition_model} below.

Our central question is: \emph{what structure must a maximally deviating subgroup possess, and how does imposing a support constraint alter that structure?} We show that both questions admit sharp answers that together characterize the full deviation-support Pareto-front and reduce an otherwise intractable search to a structured combinatorial problem over a finite set of atoms.

In the following, we describe our theoretical results by first considering subgroups without support constraints, and then moving to our main theorem in the constrained case.

\textbf{Subgroups without support constraints.} We begin with the following observation that constrains what any optimal subgroup can look like. A subgroup \emph{cannot} benefit from internal heterogeneity: if some sub-region of $Q$ had a different conditional distribution from the rest, one could split $Q$ along that boundary and keep the more deviating side, strictly improving on $Q$. The optimum must therefore be internally uniform. Formally (all proofs are deferred to Appendix~\ref{app:proofs}):

\begin{restatable}[Homogeneity of Optimal Subgroups]{theorem}{homogeneitythm}
\label{thm:homogeneity}
Let $(X,Y)\sim P$. Suppose $Q^\star \subseteq \mathcal{X}$ maximizes $d(Q)$ over all measurable $Q$ with $P(X\in Q)>0$. Then the conditional distribution of $Y$ is constant throughout $Q^\star$: for every measurable $Q' \subset Q^\star$ with $0 < P(X\in Q') < P(X\in Q^\star)$,
\[
P(Y \mid X \in Q') \;=\; P(Y \mid X \in Q^\star),
\]
and consequently $d(Q') = d(Q^\star \setminus Q') = d(Q^\star)$.
\end{restatable}

\camera{Practically, it may be more insightful to read Theorem~\ref{thm:homogeneity} in its contrapositive form, which does not require such a $Q^\star$ with positive mass to exist: if a subgroup $Q$ is not homogeneous in the above sense, i.e., there exists $Q' \subset Q$ with $P(Y \mid X \in Q') \neq P(Y \mid X \in Q \setminus Q')$, then $d(Q) < \max\{d(Q'),\, d(Q \setminus Q')\}$ and hence $Q$ must \emph{not} be the maximizer.}

\camera{Thus,} Theorem~\ref{thm:homogeneity} says the optimum is already internally uniform in a distributional sense: whatever region achieves maximum deviation, every non-trivial piece of it has the same conditional law as the whole. This is a strong internal-structure constraint that holds for \emph{any} distribution $P$, without modeling assumptions. It also motivates viewing the feature space as built out of regions of homogeneous conditional distribution, each of which behaves as an \emph{atom} for the purpose of subgroup discovery\camera{: splitting it never changes its conditional distribution}. The following definition formalizes this picture and will serve as the generative model underlying the rest of our analysis.

\begin{definition}[Partition-based Generative Model]
\label{def:partition_model}
We say $(X,Y)$ follows a \emph{partition-based model} with generating subgroups $\mathcal{S} = \{S_1, \dots, S_m\}$ if $\{S_i\}_{i\in \{1,...,m\}}$ partition $\mathcal{X}$ (i.e., $\bigcup \{S_i\}_{i\in \{1,...,m\}} = \mathcal{X}$ and all pairs $S_i \cap S_j=\emptyset$ \camera{for $i \neq j$})\camera{, each atom has positive mass $P(X\in S_i)>0$}, and $P(Y \mid X = x) = P(Y \mid X \in S_i)$ for all $x \in S_i$ and all $i$.
\end{definition}

This model class is a natural fit for the problem rather than a strong additional assumption. Every rule-based subgroup discovery method already returns a rule paired with a single probabilistic model of $Y$ given the subgroup---this is what ``a subgroup'' means operationally in the literature. Assuming the data-generating process has the same form is therefore the standard realizability assumption for this setting, analogous to assuming the true regressor lies in the hypothesis class in supervised learning. Real data will of course deviate from this model to varying degrees; our experiments in Section~\ref{sec:experiments} test whether the structural consequences we derive below remain useful when it does.

Under this model, Theorem~\ref{thm:homogeneity} has a much sharper consequence than mere internal homogeneity: the global search over all measurable subsets of $\mathcal{X}$ collapses to a search over just $m$ candidates.

\begin{restatable}[Optimality of Generating Subgroups]{theorem}{optimalitylemma}
\label{thm:optimality}
Under the partition-based model of Definition~\ref{def:partition_model},
\[
\max_{Q \subseteq \mathcal{X}} d(Q) \;=\; \max_{i \in \{1, \dots, m\}} d(S_i).
\]
\camera{Further, if the maximizer is unique, i.e., $d(S_k) > d(S_i)$ for all $i \neq k$, then the optimal subgroups are exactly $S_k$ and, by homogeneity, its subsets: a subgroup $Q$ with $P(X\in Q)>0$ attains the maximum if and only if $P(X \in Q \setminus S_k) = 0$. In particular, no subgroup that places positive mass outside $S_k$, whether another atom or any union or mixture of atoms, attains $d(S_k)$.}
\end{restatable}

The reduction here is dramatic: an uncountable search space becomes a finite one of size $m$, with no loss of optimality. Practically, this means that once a good partition of $\mathcal{X}$ is in hand, unconstrained subgroup discovery reduces to checking $m$ candidates. This yields a modular recipe: first fit a partition-based model of $P(Y \mid X)$ via standard supervised learning (e.g., a decision tree with a non-parametric estimate of the conditional distribution at each leaf), then pick the leaf whose conditional distribution has the largest KL to the marginal. Subgroup discovery, which has classically required bespoke combinatorial search, thus reduces to conditional distribution estimation followed by a trivial argmax over leaves.

\textbf{Subgroups with support constraints.} In practice with finite samples, a minimum support constraint $P(X \in Q) \geq \alpha$ is usually required, since very small subgroups are statistically unreliable (see Appendix~\ref{appendix:subsec:instability_kl} for more thorough discussions with a bootstrap experiment) or uninteresting (non-actionable). As we show next, the clean atom-level structure of Theorem~\ref{thm:optimality} carries over to this constrained setting almost intact. We first state the problem formally.

\begin{definition}[Support-Constrained Optimal Subgroup]
\label{def:support_constrained}
Given a minimum support threshold $\alpha \in (0, 1)$, define
\begin{equation}
    Q^* \;=\; \mathop{\arg\max}_{Q \subseteq \mathcal{X}:\; P(X \in Q) \geq \alpha}\; d(Q).
\end{equation}
\end{definition}

Our core theoretical contribution, as shown by the next \camera{theorem}, reveals that it is never advantageous to take partial slices from two different generating regions simultaneously. Therefore, $Q^*$ is a union of whole atoms together \textbf{with at most one fractional atom}. Specifically, Theorem~\ref{thm:support_existence} below establishes that a subgroup of this form \emph{suffices} for optimality; Corollary~\ref{cor:uniqueness} shows that under distinct generating distributions this structure is also \emph{necessary}.

\begin{restatable}[Main Result: Structure of Support-Constrained Optimal Subgroups]{theorem}{supportexistencethm}
\label{thm:support_existence}
Let $\mathcal{S} = \{S_1, \dots, S_m\}$ be the generating partition with capacities $\pi_i = P(X \in S_i)$\camera{, and assume the atoms are \emph{divisible}: for every $i$ and every $\beta \in [0, \pi_i]$, some $S' \subseteq S_i$ has $P(X \in S') = \beta$.}
For any $\alpha \in (0,1)$, there \emph{exists} an optimal solution $Q^*$ to Definition~\ref{def:support_constrained} of the form
\[
Q^* \;=\; \left(\bigcup_{i \in \mathcal{I}} S_i\right) \cup S'_k,
\]
for some index set $\mathcal{I}$ and \camera{some $S'_k \subseteq S_k$ with $k \notin \mathcal{I}$ (possibly $S'_k = \emptyset$), i.e., at most one atom is included partially.}
\end{restatable}

\begin{restatable}[Uniqueness under Distinct Distributions]{corollary}{uniquenesscor}
\label{cor:uniqueness}
\camera{Under the assumptions of Theorem~\ref{thm:support_existence}, if} the generating distributions $\{P(Y \mid X \in S_i)\}$ are distinct, then \emph{every} optimal solution must have the structure of Theorem~\ref{thm:support_existence}. In particular, no optimal subgroup can contain strictly partial amounts of two or more different generating regions.
\end{restatable}

Theorem~\ref{thm:support_existence} gives a complete discrete characterization of the deviation-support Pareto front: at every support level $\alpha$, the optimal subgroup is pinned down by which generating atoms to include in full, plus at most one fractional component. The entire Pareto front is therefore traced by a combinatorial selection over $m$ atoms, rather than a continuous search over all measurable subsets of $\mathcal{X}$.

Two observations make this reduction strong. First, $m$ \camera{is a property of the generating partition alone (at most the number of regions needed to separate the distinct conditional laws of $Y$ given $X$)}, whereas the unreduced search over rules in continuous feature spaces effectively enumerates all candidate splitting points per feature---a far larger combinatorial object. Second, the atoms depend on $P$ alone, independent of $\alpha$: a single partition of $\mathcal{X}$ determines the atom set, and the same set traces the Pareto front across every support level.

\camera{We close the theory section with a remark} on how these results resolve a longstanding design tension between tractability and expressiveness in the subgroup discovery literature.

\smallskip \begin{remark}[Representational Limits of Single-Conjunction Methods]
\label{rem:dnf}
Theorem~\ref{thm:support_existence} also sharpens a longstanding design tension in subgroup discovery. For tractability, most existing methods restrict $Q$ to a conjunction of conditions and avoid \emph{disjunctive normal form} (DNF), i.e., unions of rules, since searching over arbitrary DNFs is combinatorially expensive. Under realizability\camera{, with atoms given by rules (as in Section~\ref{sec:method})}, however, the optimal subgroup \emph{is} a DNF: a union of atomic rules. Any single-conjunction method is therefore representationally limited and will in general miss the optimum. Our atom-union formulation replaces the intractable search over arbitrary DNFs with a selection problem over a small set of $m$ learned atoms, making DNF-aware subgroup discovery practical for the first time.
\end{remark}

\section{Learning the KL-optimal subgroup given any support threshold}
\label{sec:method}
Theorem~\ref{thm:support_existence} guarantees that the support-constrained optimum has the form $(\bigcup_{i \in \camera{\mathcal{I}}} S_i) \cup S_k'$: a union of whole atoms together with at most one fractional atom. Learning such a subgroup from data therefore decomposes into two problems: estimating the atom set $\{S_i\}$, and selecting $\mathcal{I}$ and $S_k'$ under the support constraint. We address each in turn and refer to the combined framework as \emph{AtomSD}.

The theorem reduces the problem as far as structural arguments can: further simplification would require distributional assumptions beyond partition-based realizability. \camera{Even the sub-problem of selecting which whole atoms to include is NP-hard in the worst case---we give a reduction from \textsc{Subset-Sum} in Appendix~\ref{appendix:nphard}---which rules out a polynomial-time exact algorithm and motivates a principled heuristic.} We adopt a greedy algorithm operating directly within the structural form guaranteed by Theorem~\ref{thm:support_existence}.

We deliberately instantiate both steps with simple off-the-shelf components\camera{:} if a simple instantiation of the theory already outperforms existing specialized algorithms, the gains are attributable to the structural simplification our theorems provide, not to algorithmic sophistication. While our theoretical results apply to both continuous and discrete targets, the instantiation and experiments focus on the arguably more challenging case of continuous targets.

\textbf{Step 1: Learning an approximate generating partition via CDTree.}
We employ the conditional density estimation tree (CDTree) model~\citep{yang2024conditional} to induce a partition of $\mathcal{X}$ into disjoint leaf regions \camera{$\mathcal{S} = \{S_1, \dots, S_K\}$, where each $S_i$} is a conjunction of feature constraints paired with a non-parametric (histogram) estimate of $P(Y \mid X \in \camera{S_i})$. CDTree jointly optimizes the tree structure, splitting conditions, and leaf densities; model complexity is controlled via a minimum description length criterion, removing the need to tune regularization parameters~\citep{yang2024conditional}. The resulting leaves provide a data-driven approximation of the homogeneous conditional distributions required by our theory, and we treat them as \emph{atomic subgroups} throughout\camera{, reusing the notation of Section~\ref{sec:theory} (with $K$ rather than $m$ to emphasize that the leaves are learned rather than given)}.

\textbf{Step 2: Selecting atoms via Greedy Atomic Pruning (GAP).}
\camera{Following Theorem~\ref{thm:support_existence}, a candidate subgroup is described by an index set $\mathcal{I} \subseteq [K] := \{1, \dots, K\}$ of whole atoms together with an optional partial component $p \subseteq S_k$ for some $k \notin \mathcal{I}$; let $\mathrm{score}(\mathcal{I}, p)$ denote the empirical KL divergence of $(\bigcup_{i \in \mathcal{I}} S_i) \cup p$.} GAP (Algorithm~\ref{alg:expansion-union}) maintains a current solution $\camera{(\mathcal{I}, p)}$ and alternates two greedy moves until convergence. \camera{We slightly abuse terminology and also use ``support'' for its empirical counterpart, the number of training instances covered by the subgroup, and write $\alpha$ for the minimum support threshold in this sample-level sense as well.} We describe each of them next. 

\begin{algorithm}[h]
\caption{Greedy Atomic Pruning (GAP)}\label{alg:expansion-union}
\begin{algorithmic}[1]\cameracolor
\Require Candidate atomic subgroups (CDTree leaves) $\mathcal{S} = \{S_1, \dots, S_K\}$, training data $(x^n, y^n)$, minimum support $\alpha$. Throughout, $\Call{Score}{\mathcal{I}, p}$ denotes the empirical KL divergence of $(\bigcup_{i \in \mathcal{I}} S_i) \cup p$ on the training data and $\Call{supp}{\mathcal{I}, p}$ the number of training instances it covers.
\Ensure The learned subgroup, given as an index set $\mathcal{I} \subseteq [K]$ of whole atoms and an optional partial subgroup $p$.
\Statex
\State $\mathcal{I} \gets [K]$; \quad $p \gets \varnothing$ \Comment{Initialize with all candidate atoms}
\State $(\mathcal{I}, p) \gets$ \Call{SearchFull}{$\mathcal{I}, p$} \Comment{Initial greedy selection of whole atoms}
\Repeat \Comment{Alternate partial refinement and whole-atom selection}
    \State $\mathit{improved} \gets \textbf{false}$
    \State $(\mathcal{I}', p') \gets$ \Call{SearchPartial}{$\mathcal{I}, p$} \Comment{See Algorithm~\ref{alg:search partial} in Appendix~\ref{app:sec:alg}}
    \If{$\Call{Score}{\mathcal{I}', p'} > \Call{Score}{\mathcal{I}, p}$}
        \State $(\mathcal{I}, p) \gets (\mathcal{I}', p')$; \quad $\mathit{improved} \gets \textbf{true}$
    \EndIf
    \State $(\mathcal{I}', p') \gets$ \Call{SearchFull}{$\mathcal{I}, p$} \Comment{See Algorithm~\ref{alg:search full} in Appendix~\ref{app:sec:alg}}
    \If{$\Call{Score}{\mathcal{I}', p'} > \Call{Score}{\mathcal{I}, p}$}
        \State $(\mathcal{I}, p) \gets (\mathcal{I}', p')$; \quad $\mathit{improved} \gets \textbf{true}$
    \EndIf
\Until{$\lnot\, \mathit{improved}$}
\State \Return $(\mathcal{I}, p)$
\end{algorithmic}
\end{algorithm}

\emph{SearchFull} refines the \camera{index set $\mathcal{I}$} with $p$ held fixed. Starting from $\camera{\mathcal{I} = [K]}$ and $p = \varnothing$, it considers every single-atom toggle---adding an atom currently absent from $\camera{\mathcal{I}}$, or removing one currently present---and commits to the best improving move, subject to the minimum-support constraint \camera{that $(\bigcup_{i \in \mathcal{I}} S_i) \cup p$ covers at least $\alpha$ training instances}. The loop terminates when no single-atom toggle improves $\mathrm{score}$.

\emph{SearchPartial} refines the fractional component $p$ with the structure of $\camera{\mathcal{I}}$ held fixed. For each candidate atom $\camera{S_i}$, it considers using $\camera{S_i}$ as the fractional atom (removing \camera{$i$} from $\camera{\mathcal{I}}$ if present) and \qh{greedily shrinks its bounding box. At each step it considers every \emph{cut}, that is, raising the lower bound or lowering the upper bound of a single feature to a value observed within the current box, and applies the cut that most improves $\mathrm{score}$, until no cut improves it}. 
The candidate $\camera{(\mathcal{I}', p'_i)}$ achieving the highest score across all \camera{$i$} is returned.

The two moves alternate until neither improves $\mathrm{score}$. The output is a subgroup of the form prescribed by Theorem~\ref{thm:support_existence}: a union of whole atoms plus at most one shrunken partial atom. Full pseudocode for SearchFull and SearchPartial is given in Algorithms~\ref{alg:search full} and~\ref{alg:search partial} in Appendix~\ref{app:sec:alg}.

\smallskip \textbf{A second use of the theory: improving subgroups returned by other methods.}
Theorem~\ref{thm:support_existence} and Corollary~\ref{cor:uniqueness} jointly imply a structural \emph{necessity} condition: under distinct generating distributions, any Pareto-optimal subgroup must be a union of whole atoms plus at most one fractional atom. A subgroup that cuts partially through two or more atoms violates this condition and therefore cannot lie on the Pareto front; however, it can often be modified into one that does.

AtomSD exploits this observation directly. Given a subgroup $Q$ produced by any existing subgroup discovery algorithm, we restrict \camera{$\mathcal{S}$} to the CDTree leaves overlapping $Q$, initialize \camera{$\mathcal{I}$ to their indices}, and apply GAP (Algorithm~\ref{alg:expansion-union}) within this restricted candidate set. The resulting subgroup satisfies the necessity condition by construction. As we show in Section~\ref{sec:experiments}, this post-processing reliably improves subgroups produced by every competitor we test. This is \camera{evidence} that the condition is frequently violated in practice, and that enforcing it recovers otherwise unavailable KL divergence.

\section{Experiments} \label{sec:experiments}

Our experiments aim to answer whether the structural insights of Section~\ref{sec:theory} transfer to practice, i.e., when the partition must be learned from finite data and the partition-based generative assumption holds only approximately. We organize the evaluation around \camera{Theorem~\ref{thm:support_existence} and Corollary~\ref{cor:uniqueness}}: \emph{sufficiency} (Section~\ref{subsec:exp:constrained}), that a subgroup of the atom-union form suffices for Pareto-optimality; and \emph{necessity} (Section~\ref{subsec:exp:improve}), that any subgroup violating this form can be improved by enforcing it. 

\subsection{General setups}
\noindent \textbf{Datasets.} We use 19 \camera{benchmark datasets from the UCI repository~\citep{uci_link},} commonly used in the subgroup discovery literature~\citep{proencca2020discovering, yang2024conditional, xu2024learning}, spanning a wide range of sample sizes and dimensionalities (see Table~\ref{tab:data summary} in Appendix~\ref{app:sec:setup}).

\noindent \textbf{Competitor algorithms.} We compare against four methods: SyFlow~\citep{xu2024learning}, a state-of-the-art neural approach that optimizes a support-weighted KL objective via stochastic gradient descent; RSD~\citep{proencca2022robust}, a strong heuristic-search method based on the minimum description length principle; and the classic subgroup discovery methods SD-$\mu$ and SD-KL~\citep{lemmerich2018pysubgroup} for numeric targets.

\noindent \textbf{Evaluation.} All results are averaged over 3-fold cross-validation: the training set is used to obtain subgroups, and KL divergence is estimated on the held-out test set to avoid the bias of training-set evaluation. Full reproducibility details are provided in Appendix~\ref{app:sec:setup}, and our source code is available at \url{ https://github.com/ylincen/AtomSD}.

\subsection{Sufficiency: AtomSD finds deviating subgroups at every support threshold} \label{subsec:exp:constrained}

The theory predicts that the atom-union form suffices for Pareto-optimality at any support threshold. The empirical question is whether this remains true when the partition is learned from finite data rather than given exactly, and whether it translates into better subgroups than methods committed to a single implicit trade-off.

\begin{table}[ht]
\centering
\caption{Average \camera{held-out KL divergence} (higher is better) across 3-fold CV for support threshold $\alpha$. \textbf{Bold} indicates the best method per row. If a method returns no subgroup satisfying the support threshold on \camera{none} of the three folds, KL is set to 0 (the default, trivial subgroup covering all instances); otherwise, the reported value is averaged only over folds on which the method returns a valid subgroup. \camera{Held-out KL estimates can be negative; see Appendix~\ref{app:sec:setup}.}}
\label{tab:exp3}
\resizebox{0.98\textwidth}{!}{%
\begin{tabular}{llrrrrr@{\quad}llrrrrr}
\toprule
Dataset & $\alpha$ & \textbf{AtomSD} & RSD & SD-KL & SD-$\mu$ & SyFlow & Dataset & $\alpha$ & \textbf{AtomSD} & RSD & SD-KL & SD-$\mu$ & SyFlow \\
\midrule
\multirow{5}{*}{Ailerons} & 50 & \textbf{1.768} & 1.363 & 0.981 & 1.053 & 1.312 & \multirow{5}{*}{SkillCraft} & 50 & \textbf{1.909} & 1.035 & -0.458 & 0.699 & 0.592 \\
 & 100 & \textbf{1.718} & 1.244 & 0.981 & 1.053 & 1.312 &  & 100 & \textbf{1.532} & 0.000 & -0.458 & 0.699 & 0.592 \\
 & 150 & \textbf{1.684} & 0.000 & 0.981 & 1.053 & 1.312 &  & 150 & \textbf{0.715} & 0.000 & -0.458 & 0.699 & 0.592 \\
 & 200 & \textbf{1.475} & 0.000 & 0.981 & 1.053 & 1.312 &  & 200 & \textbf{0.715} & 0.000 & -0.459 & 0.699 & 0.592 \\
 & 300 & \textbf{1.684} & 0.000 & 0.981 & 1.053 & 1.312 &  & 300 & \textbf{0.715} & 0.000 & 0.000 & 0.699 & 0.645 \\
\midrule
\multirow{5}{*}{Bikesharing} & 50 & \textbf{3.679} & 0.891 & 1.290 & 2.889 & 1.246 & \multirow{5}{*}{SML2010} & 50 & \textbf{2.506} & 2.029 & 0.000 & 1.180 & 1.253 \\
 & 100 & \textbf{3.288} & 0.882 & 1.290 & 2.889 & 1.246 &  & 100 & \textbf{1.868} & 0.000 & 0.000 & 1.180 & 1.063 \\
 & 150 & \textbf{3.151} & 0.923 & 1.290 & 2.818 & 1.246 &  & 150 & \textbf{1.552} & 0.000 & 0.000 & 1.180 & 1.063 \\
 & 200 & \textbf{3.043} & 0.725 & 1.290 & 2.818 & 1.246 &  & 200 & \textbf{1.397} & 0.000 & 0.000 & 1.180 & 0.925 \\
 & 300 & 2.771 & 0.000 & 1.290 & \textbf{2.818} & 1.246 &  & 300 & \textbf{1.151} & 0.000 & 0.000 & 0.000 & 0.948 \\
\midrule
\multirow{5}{*}{California} & 50 & \textbf{3.936} & -0.694 & 0.000 & 1.316 & 0.747 & \multirow{5}{*}{Supercond} & 50 & \textbf{2.620} & 0.855 & 0.839 & 0.773 & 0.665 \\
 & 100 & \textbf{3.936} & 0.000 & 0.000 & 1.316 & 0.747 &  & 100 & \textbf{2.956} & 0.896 & 0.839 & 0.773 & 0.665 \\
 & 150 & \textbf{3.936} & 0.000 & 0.000 & 1.316 & 0.747 &  & 150 & \textbf{3.023} & 0.000 & 0.839 & 0.773 & 0.665 \\
 & 200 & \textbf{3.936} & 0.000 & 0.000 & 1.316 & 0.747 &  & 200 & \textbf{2.925} & 0.000 & 0.839 & 0.773 & 0.665 \\
 & 300 & \textbf{2.504} & 0.000 & 0.000 & 1.316 & 0.747 &  & 300 & \textbf{2.672} & 0.000 & 0.839 & 0.773 & 0.665 \\
\midrule
\multirow{5}{*}{Concrete} & 50 & \textbf{1.278} & 1.174 & 0.000 & 1.220 & 1.213 & \multirow{5}{*}{Support2} & 50 & \textbf{0.919} & 0.119 & -0.181 & -0.175 & -0.163 \\
 & 100 & \textbf{0.594} & 0.000 & 0.000 & 0.000 & 0.460 &  & 100 & \textbf{0.589} & 0.124 & -0.181 & -0.226 & -0.163 \\
 & 150 & \textbf{0.510} & 0.000 & 0.000 & 0.000 & 0.186 &  & 150 & \textbf{0.801} & 0.125 & -0.181 & -0.253 & -0.163 \\
 & 200 & \textbf{0.390} & 0.000 & 0.000 & 0.000 & 0.000 &  & 200 & \textbf{0.406} & 0.125 & -0.181 & -0.276 & -0.163 \\
 & 300 & \textbf{0.266} & 0.000 & 0.000 & 0.000 & 0.000 &  & 300 & \textbf{0.142} & 0.000 & -0.181 & 0.000 & -0.163 \\
\midrule
\multirow{5}{*}{Elevators} & 50 & 0.318 & 0.000 & -0.268 & -0.100 & \textbf{0.447} & \multirow{5}{*}{Synchronous} & 50 & 1.375 & \textbf{1.525} & 0.000 & 1.204 & 0.965 \\
 & 100 & 0.318 & 0.000 & -0.268 & -0.100 & \textbf{0.447} &  & 100 & \textbf{0.846} & 0.000 & 0.000 & 0.000 & 0.789 \\
 & 150 & 0.318 & 0.000 & -0.268 & -0.100 & \textbf{0.447} &  & 150 & \textbf{0.593} & 0.000 & 0.000 & 0.000 & 0.554 \\
 & 200 & 0.318 & 0.000 & -0.268 & -0.100 & \textbf{0.447} &  & 200 & \textbf{0.449} & 0.000 & 0.000 & 0.000 & 0.000 \\
 & 300 & 0.318 & 0.000 & -0.268 & -0.100 & \textbf{0.447} &  & 300 & \textbf{0.106} & 0.000 & 0.000 & 0.000 & 0.000 \\
\midrule
\multirow{5}{*}{Energy} & 50 & \textbf{2.272} & 2.229 & 1.088 & 1.250 & 0.844 & \multirow{5}{*}{Thermography} & 50 & \textbf{0.946} & 0.554 & 0.000 & 0.000 & 0.451 \\
 & 100 & 1.846 & \textbf{2.017} & 1.088 & 1.250 & 0.844 &  & 100 & \textbf{0.591} & 0.412 & 0.000 & 0.000 & 0.272 \\
 & 150 & \textbf{1.955} & 1.720 & 1.088 & 1.250 & 0.844 &  & 150 & \textbf{0.524} & 0.349 & 0.000 & 0.000 & 0.000 \\
 & 200 & \textbf{1.871} & 1.720 & 1.088 & 1.250 & 0.844 &  & 200 & \textbf{0.319} & 0.289 & 0.000 & 0.000 & 0.000 \\
 & 300 & \textbf{1.800} & 0.000 & 1.088 & 1.250 & 0.844 &  & 300 & \textbf{0.243} & 0.000 & 0.000 & 0.000 & 0.000 \\
\midrule
\multirow{5}{*}{ForestFires} & 50 & 0.000 & 0.131 & 0.000 & 0.000 & \textbf{0.579} & \multirow{5}{*}{Toxicity} & 50 & \textbf{0.897} & 0.465 & 0.000 & 0.000 & 0.617 \\
 & 100 & 0.000 & 0.000 & 0.000 & 0.000 & 0.000 &  & 100 & \textbf{0.682} & 0.334 & 0.000 & 0.000 & 0.544 \\
 & 150 & 0.000 & 0.000 & 0.000 & 0.000 & 0.000 &  & 150 & \textbf{0.385} & 0.000 & 0.000 & 0.000 & 0.000 \\
 & 200 & 0.000 & 0.000 & 0.000 & 0.000 & 0.000 &  & 200 & \textbf{0.250} & 0.000 & 0.000 & 0.000 & 0.000 \\
 & 300 & 0.000 & 0.000 & 0.000 & 0.000 & 0.000 &  & 300 & \textbf{0.088} & 0.000 & 0.000 & 0.000 & 0.000 \\
\midrule
\multirow{5}{*}{House} & 50 & \textbf{3.609} & 0.000 & 0.000 & 1.501 & 0.447 & \multirow{5}{*}{Treasury} & 50 & 1.605 & 1.282 & 0.000 & \textbf{1.923} & 0.817 \\
 & 100 & \textbf{3.141} & 0.000 & 0.000 & 1.468 & 0.447 &  & 100 & \textbf{1.302} & 0.000 & 0.000 & 0.000 & 0.817 \\
 & 150 & \textbf{2.357} & 0.000 & 0.000 & 0.000 & 0.447 &  & 150 & \textbf{1.184} & 0.000 & 0.000 & 0.000 & 0.817 \\
 & 200 & \textbf{1.559} & 0.000 & 0.000 & 0.000 & 0.447 &  & 200 & \textbf{1.006} & 0.000 & 0.000 & 0.000 & 0.817 \\
 & 300 & \textbf{1.495} & 0.000 & 0.000 & 0.000 & 0.447 &  & 300 & \textbf{0.782} & 0.000 & 0.000 & 0.000 & 0.607 \\
\midrule
\multirow{5}{*}{NavalProp} & 50 & \textbf{1.525} & 1.242 & 0.878 & 0.449 & 0.952 & \multirow{5}{*}{Wizmir} & 50 & 1.516 & \textbf{1.801} & 1.629 & 1.072 & 1.016 \\
 & 100 & \textbf{1.649} & 1.420 & 0.878 & 0.449 & 0.952 &  & 100 & 1.534 & 1.226 & \textbf{1.629} & 1.072 & 1.016 \\
 & 150 & \textbf{1.693} & 0.000 & 0.494 & 0.449 & 0.813 &  & 150 & \textbf{1.429} & 0.000 & 0.000 & 1.072 & 1.016 \\
 & 200 & \textbf{1.681} & 0.000 & 0.494 & 0.120 & 0.448 &  & 200 & \textbf{1.115} & 0.000 & 0.000 & 1.072 & 1.016 \\
 & 300 & \textbf{1.635} & 0.000 & 0.494 & 0.120 & 0.003 &  & 300 & 0.839 & 0.000 & 0.000 & 0.000 & \textbf{0.920} \\
\midrule
\multirow{5}{*}{Puma32h} & 50 & \textbf{3.174} & 3.088 & 0.574 & 1.913 & 0.439 &  &  &  &  &  &  &  \\
 & 100 & \textbf{2.686} & 1.885 & 0.574 & 1.913 & 0.439 &  &  &  &  &  &  &  \\
 & 150 & \textbf{2.640} & 1.312 & 0.574 & 1.913 & 0.439 &  &  &  &  &  &  &  \\
 & 200 & \textbf{2.549} & 1.361 & 0.574 & 1.913 & 0.439 &  &  &  &  &  &  &  \\
 & 300 & \textbf{2.335} & 0.000 & 0.574 & 1.913 & 0.439 &  &  &  &  &  &  &  \\
\midrule
\#Best & 50 & \textbf{14} & 2 & 0 & 1 & 2 & Avg.\ Rank & 50 & \textbf{1.42} & 2.61 & 4.29 & 3.32 & 3.37 \\
 & 100 & \textbf{15} & 1 & 1 & 0 & 1 &  & 100 & \textbf{1.26} & 3.24 & 3.97 & 3.42 & 3.11 \\
 & 150 & \textbf{17} & 0 & 0 & 0 & 1 &  & 150 & \textbf{1.16} & 3.79 & 3.89 & 3.18 & 2.97 \\
 & 200 & \textbf{17} & 0 & 0 & 0 & 1 &  & 200 & \textbf{1.16} & 3.74 & 3.79 & 3.13 & 3.18 \\
 & 300 & \textbf{15} & 0 & 0 & 1 & 2 &  & 300 & \textbf{1.26} & 4.08 & 3.63 & 3.03 & 3.00 \\
\textit{Total} & -- & \textbf{78} & 3 & 1 & 2 & 7 & \textit{Overall} & -- & \textbf{1.25} & 3.49 & 3.92 & 3.22 & 3.13 \\
\bottomrule
\end{tabular}}
\end{table}

Table~\ref{tab:exp3} shows a clear pattern. As the minimum support $\alpha$ increases, every competitor increasingly fails to return any subgroup satisfying the threshold; in this case, we fall back to treating the whole dataset as the trivial subgroup, which by construction yields KL~$=0$. \camera{By $\alpha \geq 150$, RSD returns no valid subgroup on most datasets, SD-KL on about half, and SD-$\mu$ and SyFlow on a growing number of them.} This is exactly the failure mode \camera{anticipated in} the introduction: these methods optimize a folded quality measure that commits to one point on the Pareto front, and when the user's support requirement disagrees with that point, the method has nothing to fall back on. In principle one could re-tune each competitor's hyperparameters per threshold, but this requires knowledge of each algorithm's internal trade-off and defeats the purpose of a support-constrained interface.

AtomSD, in contrast, returns a non-trivial subgroup on nearly every dataset at every threshold, and the KL divergences remain substantial even under the tightest constraints. In aggregate, AtomSD achieves an average rank of 1.25 and \camera{78} wins out of 95 (dataset, $\alpha$) combinations, and the pattern is consistent across datasets of varying sample size and dimensionality. A Friedman test \camera{(with the Iman--Davenport correction~\citep{demvsar2006statistical})} rejects the null of equal performance at every support level ($p < 10^{-5}$ in each case); the corresponding critical-difference diagram~\citep{demvsar2006statistical} is in Appendix~\ref{appendix:cd-diagram}. The gap is most pronounced on larger datasets such as \textit{California}, \textit{House}, \camera{and \textit{Supercond}}, where AtomSD sustains strong divergence at every $\alpha$ while \camera{most} competitors degenerate.

We read this as direct evidence that the structural insight of Theorem~\ref{thm:support_existence} transfers to the finite-sample setting: even though the partition is estimated rather than given, and even though real data deviates from the partition-based model to varying degrees, targeting the atom-union form yields Pareto-optimal subgroups in practice. \camera{Finally,} in some applications the support threshold is naturally expressed as a ratio rather than an absolute count; AtomSD performs equally strongly in that regime, as reported in Appendix~\ref{appendix:ratio_constrain_subgroup_discovery}.

\subsection{Necessity: enforcing the structural condition improves any subgroup} \label{subsec:exp:improve}
Corollary~\ref{cor:uniqueness} implies a necessity condition: under distinct generating distributions, any Pareto-optimal subgroup \emph{must} be a union of whole atoms plus at most one fractional atom. A subgroup that cuts partially through two or more atoms therefore cannot lie on the Pareto front, and can in principle be improved by enforcing the condition. We of course do not have access to the true atoms; we test instead whether enforcing the condition against the \emph{learned} atoms still improves the subgroup, which would indicate that the necessity insight transfers to practice.

\begin{figure}[ht]
    \centering
    \includegraphics[width=0.8\linewidth]{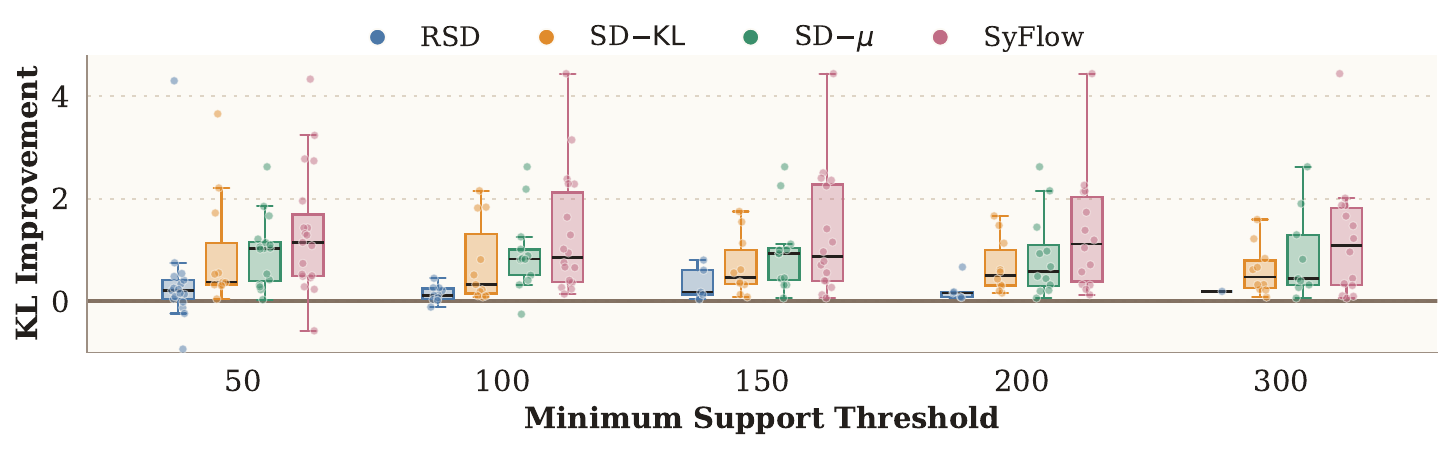}
    \caption{Held-out KL-divergence difference between AtomSD's refined subgroup and each competitor's original subgroup, across 19 datasets and five support thresholds. Positive values indicate that AtomSD's refinement improves the given subgroups.}
    \label{fig:exp2_fixed}
\end{figure}

For each competitor's returned subgroup $Q$, we apply AtomSD as a post-processing step: restrict the candidate atoms to the CDTree leaves overlapping $Q$, and re-run GAP within that restricted footprint. Figure~\ref{fig:exp2_fixed} reports the held-out KL-divergence difference between the refined subgroup and the original. Across all four competitors and every support threshold, the distribution is overwhelmingly positive: medians are consistently above zero, and negative cases are rare and \camera{mostly} small in magnitude. This pattern holds regardless of which baseline produced the initial subgroup, which we interpret as evidence that 1) the necessity condition is frequently violated in practice, and 2) enforcing it recovers additional KL divergence beyond what the original algorithm captured. The corresponding results when the support threshold is expressed as a ratio are reported in Appendix~\ref{appendix:subgroup_ratio_improvement} and show the same pattern.

Beyond the theoretical validation, this post-processing use of AtomSD has practical value: in applications where a domain expert or prior analysis has identified a candidate subgroup of interest, AtomSD can refine it into a Pareto-optimal subgroup \camera{built from the atoms overlapping the original region}.

\section{Conclusion} \label{sec:conclusion}
We revisited subgroup discovery from a probabilistic modeling perspective and gave a structural characterization of KL-optimal subgroups under support constraints. Under a partition-based generative model, the entire deviation-support Pareto front is traced by unions of whole atoms together with at most one fractional atom, collapsing an uncountable search over measurable subsets to a finite combinatorial problem. This structural reduction is {both} 1) sufficient as an optimal subgroup of this form always exists at every support level, and 2) necessary (under distinct generating distributions), as no optimal subgroup can have any other form. Together, these results overcome what has been a longstanding challenge in the literature between committing to an implicit support--deviation trade-off via a weighted scalar measure and explicitly searching under user-specified support constraints. Our experiments verify that this structural insight continues to guide useful subgroup discovery even when the partition is estimated from finite data and the generative assumption holds only approximately.

\textbf{Limitations and future work.}
Our theoretical guarantees hold under a partition-based generative model that only approximates real data; our experiments indicate that the resulting recipe remains useful when this assumption is approximately satisfied, but formal robustness results to distributional misspecification remain open. Further, future directions may include extending our theoretical results to learning subgroup sets in which the diversity among subgroups is considered.


\begin{ack}
We thank Jilles Vreeken for helpful discussions. This research was partly supported by the Dutch Research Council (NWO) through the XAIPre project (project number 19455, \url{https://doi.org/10.61686/XDODD77262}).
\end{ack}

\newpage 
\bibliography{References.bib}

\newpage
\appendix

\section{Proofs of main theorems} \label{app:proofs}
\homogeneitythm*

\begin{proof}
Fix a subset $Q$ with $P(X\in Q)>0$. Let $A \subset Q$ be a measurable subset with $0<P(X\in A)<P(X\in Q)$. Define
\[
\lambda = P(X\in A \mid X\in Q) \in (0,1),
\]
\[
P_A = P(Y\mid X\in A),
\]
\[
P_{A^c} = P(Y\mid X\in Q\setminus A).
\]
Then the conditional distribution within $Q$ decomposes as
\[
P_Q \;=\; P(Y\mid X\in Q) \;=\; \lambda P_A + (1-\lambda) P_{A^c}.
\]

By convexity of the KL divergence in its first argument,
\begin{align*}
 \mathrm{KL}(P_Q \,\|\, P(Y)) \;\le\; & \lambda \mathrm{KL}(P_A \,\|\, P(Y)) \\ 
 & + (1-\lambda) \mathrm{KL}(P_{A^c} \,\|\, P(Y)),
\end{align*}

with equality if and only if $P_A=P_{A^c}=P_Q$. 

That is,
\[
d(Q) \;\le\; \lambda d(A) + (1-\lambda)d(Q\setminus A) \;\le\; \max\{d(A),\,d(Q\setminus A)\}.
\]

Now let $Q=Q^\star$ be a global maximizer. Since $d(Q^\star)\ge d(A)$ and $d(Q^\star)\ge d(Q^\star\setminus A)$, the inequality above forces equality throughout. Thus,
\[
d(A) = d(Q^\star\setminus A) = d(Q^\star),
\]
and moreover $P_A=P_{A^c}=P_{Q^\star}$, which shows that $P(Y\mid X\in Q^\star)$ is homogeneous across all measurable subsets of $Q^\star$ with positive probability mass.
\end{proof}

\optimalitylemma*
\begin{proof}
\camera{Write $P_i := P(Y \mid X \in S_i)$ for the generating distribution of atom $S_i$. Let $Q$ be any subgroup with $P(X \in Q) > 0$, and let $w_i = P(X \in S_i \mid X \in Q)$ be the proportion of $Q$ that falls into $S_i$, so that $w_i \ge 0$ and $\sum_{i=1}^m w_i = 1$. Under the partition-based model,
\[
P_Q := P(Y \mid X \in Q) = \sum_{i=1}^m w_i P_i.
\]
Since $\mathrm{KL}(\cdot \,\|\, P(Y))$ is convex in its first argument, Jensen's inequality gives
\begin{equation}\label{eq:opt-jensen}
d(Q) \;=\; \mathrm{KL}\Big(\sum_{i=1}^m w_i P_i \,\Big\|\, P(Y)\Big) \;\le\; \sum_{i=1}^m w_i\, d(S_i) \;\le\; \max_{i} d(S_i).
\end{equation}
Each $S_i$ is itself a feasible subgroup ($w_i = 1$), so the upper bound is attained and $\max_{Q} d(Q) = \max_i d(S_i)$.

Now suppose the maximizer $k$ is unique, i.e., $d(S_i) < d(S_k)$ for all $i \neq k$. If $P(X \in Q \setminus S_k) > 0$, then $w_k < 1$, so some $j \neq k$ has $w_j > 0$, and the second inequality in~\eqref{eq:opt-jensen} is strict:
\[
\sum_{i} w_i\, d(S_i) \;=\; d(S_k) - \sum_{i \neq k} w_i \big(d(S_k) - d(S_i)\big) \;<\; d(S_k).
\]
Hence $d(Q) < d(S_k)$, and $Q$ is not optimal. Conversely, if $P(X \in Q \setminus S_k) = 0$, then $w_k = 1$, so $P_Q = P_k$ and $d(Q) = d(S_k)$. Therefore $Q$ is optimal if and only if $P(X \in Q \setminus S_k) = 0$.}
\end{proof}

\supportexistencethm*
\begin{proof}
\camera{Under the partition-based model, $d(Q)$ depends on $Q$ only through its allocation $\big(P(X \in Q \cap S_i)\big)_{i=1}^m$, since $P(Y \mid X \in Q)$ is the mixture of the $P_i$ with weights proportional to these masses. Since the atoms are divisible, every allocation with $x_i \in [0, \pi_i]$ is realized by some subgroup, so we may optimize over allocations directly. An optimal allocation exists: the objective is continuous (and finite, by positivity of the atoms) on the compact feasible set $\{\mathbf{x} \in \prod_i [0,\pi_i] : \sum_i x_i \ge \alpha\}$.}
Let $\mathbf{x}^* = (x_1^*, \dots, x_m^*)$ be an optimal allocation vector for Definition~\ref{def:support_constrained}, where $x_i^* = P(X \in Q^* \cap S_i) \in [0, \pi_i]$ with $\pi_i = P(X \in S_i)$. We show that there exists an optimal allocation in which at most one coordinate satisfies $0 < x_i < \pi_i$ strictly; this coordinate corresponds to the fractional atom $S'_k$, and the remaining $x_i \in \{0, \pi_i\}$ determine the index set $\mathcal{I}$.

Suppose \camera{that} some optimal $\mathbf{x}^*$ has two distinct indices $j, k$ with $0 < x_j^* < \pi_j$ and $0 < x_k^* < \pi_k$. We construct a perturbation along the $(j, k)$ direction and show that either a strictly better feasible point exists (contradicting optimality) or a boundary point attains the same objective (reducing the number of strictly partial atoms by one).

Consider the one-parameter family $\mathbf{x}(\epsilon)$ obtained from $\mathbf{x}^*$ by setting $x_j(\epsilon) = x_j^* + \epsilon$ and $x_k(\epsilon) = x_k^* - \epsilon$, holding all other coordinates fixed. Since $\sum_i x_i(\epsilon) = \sum_i x_i^*$ is preserved, the support constraint $P(X \in Q) \geq \alpha$ remains satisfied for any $\epsilon$ such that $x_j(\epsilon) \in [0, \pi_j]$ and $x_k(\epsilon) \in [0, \pi_k]$. Let $[\epsilon_{\min}, \epsilon_{\max}]$ denote this feasible interval, which contains $\epsilon = 0$ in its interior by assumption.

Let $g(\epsilon) = d(Q(\mathbf{x}(\epsilon)))$ denote the objective along this segment. Under the partition-based model, $P(Y \mid X \in Q(\mathbf{x}(\epsilon)))$ is a mixture $\sum_i w_i(\epsilon) P_i$ where $w_i(\epsilon) = x_i(\epsilon)/\sum_i x_i(\epsilon)$. Since $\sum_i x_i(\epsilon)$ is constant in $\epsilon$, the mixture weights $w_i(\epsilon)$ are affine in $\epsilon$. The KL divergence is convex in its first argument, so $g(\epsilon)$ is a convex function of $\epsilon$.

\textbf{Case 1 ($P_j \neq P_k$).} The mixture distribution $\sum_i w_i(\epsilon) P_i$ changes non-trivially with $\epsilon$ (since only $w_j$ and $w_k$ vary, and $P_j \neq P_k$), so $g$ is \emph{strictly} convex on $[\epsilon_{\min}, \epsilon_{\max}]$\camera{ (all values are finite, since $dP_i/dP(Y) \le 1/\pi_i$ by positivity of the atoms)}. The maximum of a strictly convex function on a closed interval is attained at one of the endpoints, and not at any interior point. Since $\epsilon = 0$ is interior, $g(\epsilon_{\min})$ or $g(\epsilon_{\max})$ strictly exceeds $g(0)$, contradicting the optimality of $\mathbf{x}^*$.

\textbf{Case 2 ($P_j = P_k$).} The mixture $\sum_i w_i(\epsilon) P_i$ is invariant to shifting mass between $j$ and $k$, since $(w_j(\epsilon) + w_k(\epsilon)) P_j = (w_j(\epsilon) + w_k(\epsilon)) P_k$ equals $w_j(0) P_j + w_k(0) P_k$ at every $\epsilon$. Hence $g(\epsilon)$ is constant on $[\epsilon_{\min}, \epsilon_{\max}]$. In particular, $g(\epsilon_{\min}) = g(\epsilon_{\max}) = g(0)$, so we may replace $\mathbf{x}^*$ with the allocation at either endpoint without changing the objective. At an endpoint, at least one of $x_j, x_k$ equals $0$ or \camera{its capacity ($\pi_j$ or $\pi_k$, respectively)}, reducing the number of strictly partial coordinates by one.

Combining the two cases: any optimal $\mathbf{x}^*$ with two or more strictly partial coordinates can be replaced with an equally-optimal allocation having strictly fewer such coordinates. Iterating this reduction terminates with an optimal allocation $\tilde{\mathbf{x}}$ in which at most one coordinate is strictly partial. Setting $\mathcal{I} = \{i : \tilde{x}_i = \pi_i\}$, $k$ equal to the unique strictly-partial index (if any) with $S'_k \subseteq S_k$ the corresponding partial inclusion, and $S'_k = \emptyset$ otherwise, yields a solution of the form claimed in the theorem statement.
\end{proof}

\uniquenesscor*
\begin{proof}
\camera{Let $Q$ be optimal with allocation $\mathbf{x}^*$, and suppose two indices $j \neq k$ satisfy $0 < x_j^* < \pi_j$ and $0 < x_k^* < \pi_k$. Since the generating distributions are distinct, $P_j \neq P_k$, so Case~1 of the proof of Theorem~\ref{thm:support_existence} applies: $g$ is strictly convex on $[\epsilon_{\min}, \epsilon_{\max}]$ with $0$ in the interior, and one endpoint allocation has a strictly larger objective. That allocation is feasible and, by divisibility, realized by a subgroup, contradicting the optimality of $Q$. Hence at most one atom is strictly partial in $Q$: up to a set of probability zero, $Q = \big(\bigcup_{i \in \mathcal{I}} S_i\big) \cup S'_k$ with $\mathcal{I} = \{i : x^*_i = \pi_i\}$.}
\end{proof}




\section{Additional Details of Experimental Setup}
\label{app:sec:setup} 
\textbf{Datasets and implementation details.}
The processed datasets are summarized in Table~\ref{tab:data summary}. In total, we use 19 real-world datasets drawn from diverse application domains and commonly used in existing literature~\citep{proencca2020discovering},~\citep{yang2024conditional}, or~\citep{xu2024learning}. All nominal features are one-hot encoded. We benchmark our method against four competitor algorithms, with all results obtained on the held-out folds using three-fold cross-validation. For the competing methods, we use the default hyperparameter settings provided in their original implementations to ensure a fair comparison without over-tuning, while our method requires no hyperparameter tuning. All experiments are conducted on a cluster equipped with AMD EPYC 7662 64-Core Processors.
\begin{table}[ht]
\centering
\caption{Summary of the \camera{datasets} used in the experiments}
\label{tab:data summary}
\begin{tabular}{lrrr}
\toprule
Dataset & Sample Sizes & \#Discrete Features & \#Continuous Features \\
\midrule
Ailerons & 13750 & 4 & 36 \\
Bikesharing & 17379 & 13 & 4 \\
California & 20640 & 5 & 3 \\
Concrete & 1030 & 0 & 8 \\
Elevators & 16599 & 4 & 14 \\
Energy & 9568 & 0 & 4 \\
ForestFires & 517 & 0 & 12 \\
House & 22784 & 1 & 15 \\
NavalProp & 11934 & 0 & 16 \\
Puma32h & 8192 & 0 & 32 \\
SkillCraft & 3338 & 0 & 18 \\
SML2010 & 2764 & 0 & 20 \\
Supercond & 21263 & 0 & 81 \\
Support2 & 9105 & 0 & 26 \\
Synchronous & 557 & 0 & 4 \\
Thermography & 1003 & 0 & 25 \\
Toxicity & 908 & 0 & 6 \\
Treasury & 1049 & 0 & 15 \\
Wizmir & 1461 & 0 & 9 \\
\bottomrule
\end{tabular}
\end{table}

\textbf{Estimating KL for held-out test datasets.}
Similar to SyFlow~\citep{xu2024learning}, we evaluate all discovered subgroups by estimating the KL divergence between the subgroup's conditional target distribution and the overall target distribution using histogram densities on the held-out test set. Specifically, bin edges are determined from training data via the Freedman-Diaconis rule~\citep{freedman1981histogram}, with separate bin grids for the subgroup and overall distributions. Both grids are clamped to the global target range $[y_{\min}, y_{\max}]$ to increase the stability especially for small sample size cases. The overall histogram uses Laplace smoothing (pseudo-count of 1 per bin), while the subgroup histogram uses a Bayesian Dirichlet prior whose concentration parameters are proportional to the overall bin probabilities, with total prior strength equal to the number of bins. This provides data-driven shrinkage toward the overall distribution for bins with few subgroup observations. Both smoothed histograms are converted to densities by dividing by total count and bin width. The KL divergence is then estimated via a sample-wise approximation\camera{, i.e., the average of $\log \hat{p}(y) - \log \hat{q}(y)$ over the held-out subgroup members, where $\hat{p}$ and $\hat{q}$ are the subgroup and overall densities fitted on the training split. Because the densities are fitted on training data and evaluated on held-out points, this estimate can be negative (see, e.g., RSD on California in Table~\ref{tab:exp3}); a negative value indicates a subgroup whose training-set deviation does not generalize to the held-out split}.

All bin edges and density parameters are learned from training data; at test time only bin assignments are computed. This train/test separation eliminates the optimistic bias that arises when the same data is used for both subgroup selection and evaluation. However, while the bias is controlled, the \emph{variance} of the KL estimate remains high for small subgroups: as we demonstrate in Appendix~\ref{appendix:subsec:instability_kl}, bootstrap resampling reveals that subgroups with fewer than $50$ training points yield KL estimates with prohibitively large variance across replicates. 

\section{Additional Experiments}
\subsection{Support-weighted KL-divergence leads to finding spurious subgroups} \label{appendix:sec:exp_intro}
We provide the experiment details for the reproducibility of Figure~\ref{fig:rankflip} in Section~\ref{sec:intro}.
  \paragraph{Distributions.}
  The background distribution \camera{$P_0$} is a 5-component Gaussian mixture with equal weights and modes at $\{-3, -1.5, 0, 1.5, 3\}$, each with standard deviation $\sigma = 0.3$.
  We define two subgroups:
  \begin{itemize}[nosep]
      \item \textbf{Subgroup A (true):} A 5-component Gaussian mixture sharing the same modes and standard deviations as the background, but with weights $(0.35, 0.05, 0.20, 0.05, 0.35)$ that over-represent
  the extreme modes. This creates a genuine distributional shift ($\mathrm{KL}(P_A \| \camera{P_0}) > 0$). Its population support is $s_A = 0.08$.
      \item \textbf{Subgroup B (spurious):} Drawn from the background distribution itself, so $\mathrm{KL}(P_B \| \camera{P_0}) = 0$ by construction. Its population support is $s_B = 0.15$.
  \end{itemize}

  \paragraph{Procedure.}
  For each subgroup sample size $k_A \in \{10, 20, \ldots, 80\} \cup \{100\}$, we set $k_B = \lceil k_A \cdot s_B / s_A \rceil$ to reflect the proportionally larger sample size of the higher-support subgroup.
  We repeat the following 300 times independently:
  \begin{enumerate}[nosep]
      \item Draw $k_A$ i.i.d.\ samples from subgroup~A's distribution and $k_B$ i.i.d.\ samples from subgroup~B's (i.e., the background).
      \item Estimate the KL divergence of each subgroup against the background using histogram density estimation with Freedman--Diaconis binning and Laplace smoothing. The background density \camera{$P_0$} is estimated
  once from a fixed reference sample of $10{,}000$ points; both subgroup densities are estimated on the same bin grid.
      \item Compute the support-weighted score for each subgroup: $\mathrm{score} = s \times \widehat{\mathrm{KL}}$.
      \item Record whether subgroup~B incorrectly outranks subgroup~A, i.e., $\mathrm{score}_B > \mathrm{score}_A$.
  \end{enumerate}

  \paragraph{Results.} The result is shown in Figure~\ref{fig:rankflip}.
  The left panel displays the true conditional densities of subgroup~A versus the background (which coincides with subgroup~B).
  The middle panel plots the mean and 95\% quantile band of the support-weighted KL estimates as a function of $k_A$.
  The right panel reports the ranking error rate---the fraction of repetitions in which the spurious subgroup~B's score exceeds that of the true subgroup~A.
  At small sample sizes ($k_A \leq 30$), the error rate exceeds 20\%, showing that KL estimation noise combined with the higher support of the spurious subgroup leads to frequent mis-rankings.

\subsection{Instability of the estimated KL} \label{appendix:subsec:instability_kl}
To assess the reliability of KL divergence estimates for subgroups of varying sizes, we perform a bootstrap stability analysis on all subgroups discovered by five algorithms (\camera{AtomSD}, RSD, SD-KL, SD-$\mu$, SyFlow) across all datasets and cross-validation folds.

For each subgroup, we run $B = 300$ bootstrap replicates. In each replicate, we resample both the training and test sets independently with replacement, preserving the original train/test split ratio. Subgroup membership is inherited from the original indices: if a resampled row corresponds to an original subgroup member, it is included in the bootstrap subgroup (possibly with duplicates).
  We then re-estimate \camera{the KL divergence between the subgroup's target distribution $P$ and the overall target distribution $P(Y)$} using the same histogram-based procedure employed during subgroup discovery:
  bin edges are determined separately for the subgroup and overall distributions via the Freedman--Diaconis rule, with a shared global range $[\min(Y) - \camera{\delta},\, \max(Y) + \camera{\delta}]$ \camera{for a small margin $\delta$};
  the overall histogram receives Laplace smoothing ($+1$ per bin), while the subgroup histogram receives a Bayesian correction using the overall histogram as a Dirichlet prior with total pseudo-count equal to
  the number of bins;
  KL divergence is estimated by the sample-wise Monte Carlo average $\widehat{\mathrm{KL}} = \frac{1}{\camera{n_Q}} \sum_{i=1}^{\camera{n_Q}} [\log \hat{p}(y_i) - \log \hat{q}(y_i)]$, where \camera{$\hat{p}$ and $\hat{q}$ are the smoothed subgroup and overall histogram densities and} the \camera{$n_Q$} evaluation points $\{y_i\}$ are the
  subgroup members in the current split.

  For each subgroup we record the standard deviation of $\widehat{\mathrm{KL}}$ across the 300 bootstrap replicates, computed separately for the training and test splits.
  Subgroups are grouped into five size buckets by training set membership: $<$50, 50--100, 100--200, 200--300, and 300+.

Notably, evaluating $\widehat{\mathrm{KL}}$ on a held-out test set does address the \emph{bias} of the training-set estimate: because the density $\hat{p}$ is fitted on training data and evaluated on independent test
  points, the optimistic bias from overfitting the histogram to the same samples is removed.
  However, the train/test split does not reduce the \emph{variance} caused by small sample sizes.
  
\begin{figure}[h!]
    \centering
    \includegraphics[width=\linewidth]{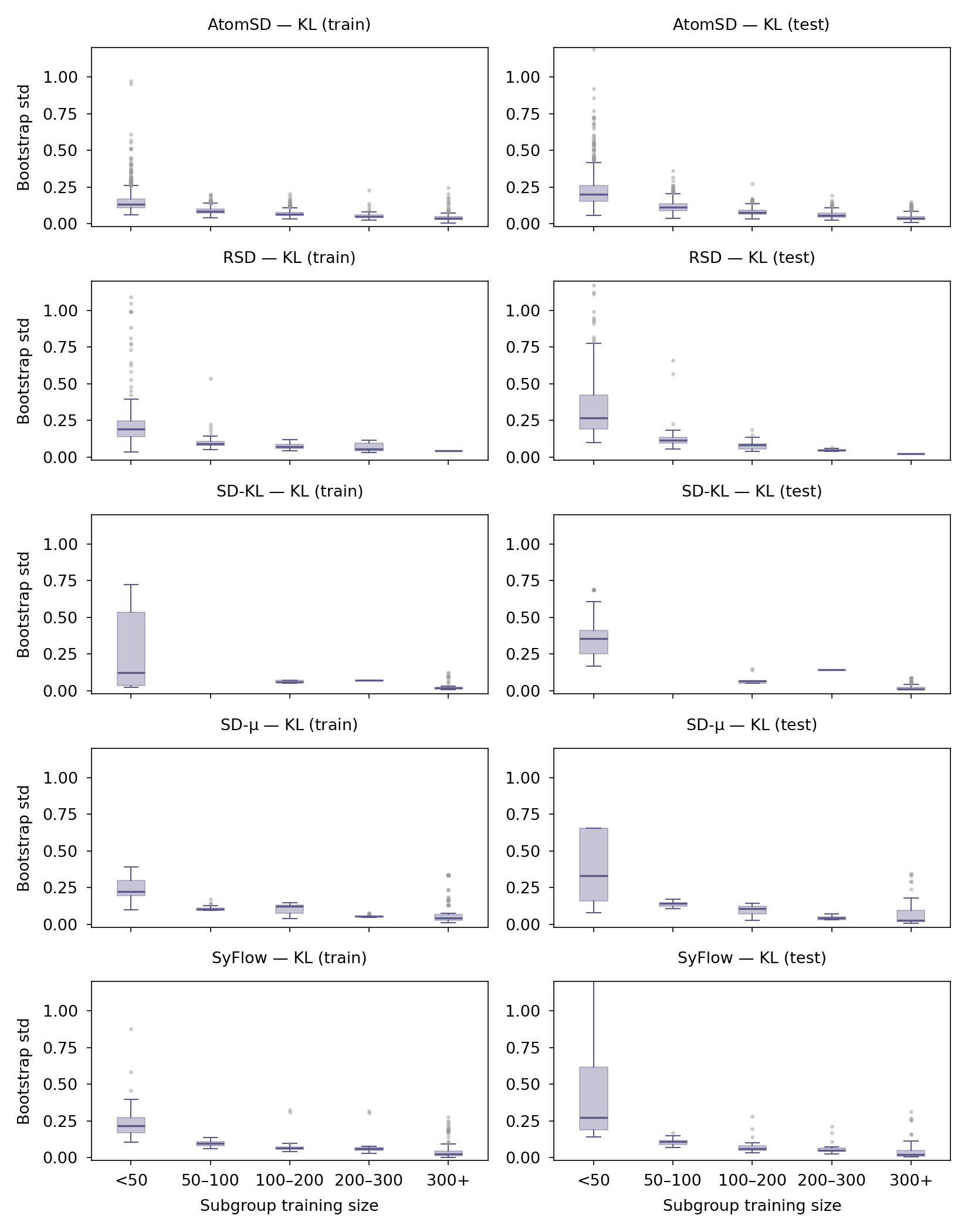}
    \caption{Bootstrap stability of KL divergence estimates across subgroup discovery algorithms. For each discovered subgroup, we perform 300 bootstrap replicates of the dataset and re-estimate
  $\widehat{\mathrm{KL}}$ using the same histogram-based procedure (Freedman--Diaconis binning with Bayesian smoothing). Each point in the box plot represents one subgroup's bootstrap standard deviation of
  $\widehat{\mathrm{KL}}$, grouped by training set size. \textbf{Left column:} KL estimated on the training split. \textbf{Right column:} KL estimated on the test split. Across all five algorithms, subgroups
  with fewer than 50 training instances exhibit substantially higher estimation variability, indicating that their KL-based ranking scores are unreliable. The pattern is consistent for both train and test
  estimates.}
      \label{fig:bootstrap-stability}
\end{figure}

  As shown in Figure~\ref{fig:bootstrap-stability}, the bootstrap standard deviation of $\widehat{\mathrm{KL}}_{\mathrm{test}}$ is comparable to---and in several cases larger than---that of
  $\widehat{\mathrm{KL}}_{\mathrm{train}}$ for small subgroups.
  This is expected: for a subgroup with $n$ training members, the histogram density $\hat{p}$ is estimated from only $n$ points regardless of the evaluation split, and the test subgroup is typically even smaller
   than the training subgroup, introducing additional Monte Carlo noise into the KL evaluation.
  In other words, the train/test split corrects \emph{what} is being estimated (removing the positive bias from evaluating on the training data), but it cannot stabilize the estimate when the underlying density
  is poorly determined due to insufficient data.

\subsection{Constrained Subgroup Discovery with support thresholds given as ratios} \label{appendix:ratio_constrain_subgroup_discovery}

When the support thresholds are given as ratios, our method shows strong performance as well, as shown in Table~\ref{table:results_exp3_ratio}.

\textbf{Support thresholds: absolute vs.\ ratio.} We report results under two conventions for the support threshold. In the main text (Table~\ref{tab:exp3}), we use absolute counts $\alpha \in \{50, 100, 150, 200, 300\}$, treating the threshold as a fixed minimum number of training instances regardless of dataset size. In the appendix here, we use ratios $\alpha' \in \{0.05, 0.10, 0.20, 0.30\}$, treating the threshold as a fraction of the training set. The two conventions probe different use cases: absolute counts correspond to an analyst requiring a minimum sample size for downstream analysis (e.g., a clinical-trial power requirement), while ratios correspond to an analyst requiring a minimum fractional coverage of the population. For a dataset with $N$ training instances, an absolute threshold of $\alpha$ corresponds to a ratio of $\alpha' = \alpha / N$, and the two conventions therefore impose identical constraints at a per-dataset level when $\alpha = \alpha' N$. Because $N$ varies across the 19 datasets by more than an order of magnitude (see Table~\ref{tab:data summary}), a fixed absolute count $\alpha$ maps to very different ratios across datasets, and vice versa; the two regimes are not directly interchangeable. 

As Appendix~\ref{appendix:subsec:instability_kl} suggests that subgroups with less than 50 training instances will lead to estimated KL-divergence with prohibitively large variance, \camera{we raise the threshold to $\max(\alpha' N, 50)$ training instances in this comparison; this matters for datasets with small sample sizes together with small $\alpha'$, whose rows then coincide with the $\alpha = 50$ rows of Table~\ref{tab:exp3}.} 

\textbf{Results.} Table~\ref{table:results_exp3_ratio} reports results under support thresholds expressed as ratios of the dataset size ($\alpha' \in \{0.05, 0.10, 0.20, 0.30\}$), complementing the absolute-count thresholds in the main text. The overall conclusion is unchanged: our method dominates, winning \camera{62} of 76 (dataset, $\alpha'$) cells and achieving the best average rank at every threshold (1.42, 1.47, 1.11, 1.26 across the four ratios, versus \camera{2.47}--4.00 for all competitors).

The gap widens as the support constraint increases. At $\alpha' = 0.20$ and $\alpha' = 0.30$, \camera{RSD, SD-KL and SD-$\mu$} collapse to the whole-dataset default (KL $= 0$) on the majority of datasets. Specifically, RSD returns no subgroup meeting the constraint on \camera{all but one dataset (Thermography)} at $\alpha' = 0.20$ or $0.30$, and SD-KL and SD-$\mu$ similarly degenerate on most large datasets. Our method, by contrast, returns a non-trivial subgroup on every dataset \camera{except ForestFires} at every threshold, with KL-divergence values that remain substantial even at $\alpha' = 0.30$ (e.g., 0.871 on NavalProp, 0.798 on Supercond, 0.760 on Puma32h). This pattern mirrors the absolute-threshold results and reflects the same underlying property: the structure theorem guarantees that a Pareto-optimal subgroup exists at every support level, and our method locates it rather than falling back to the trivial solution.

\begin{table}[t]
\centering
\caption{Average \camera{held-out KL divergence} (higher is better) across 3-fold CV for support ratio threshold $\alpha'$. \textbf{Bold} indicates the best method per row. If a method returns no subgroup satisfying the support threshold on \camera{none} of the three folds, KL is set to 0 (the default, trivial subgroup covering all instances); otherwise, the reported value is averaged only over folds on which the method returns a valid subgroup.}
\label{table:results_exp3_ratio}
\resizebox{\textwidth}{!}{%
\begin{tabular}{llrrrrr@{\quad}llrrrrr}
\toprule
Dataset & $\alpha'$ & \textbf{AtomSD} & RSD & SD-KL & SD-$\mu$ & SyFlow & Dataset & $\alpha'$ & \textbf{AtomSD} & RSD & SD-KL & SD-$\mu$ & SyFlow \\
\midrule
\multirow{4}{*}{Ailerons} & 0.05 & \textbf{1.491} & 0.000 & 0.981 & 1.053 & 1.312 & \multirow{4}{*}{SkillCraft} & 0.05 & \textbf{0.759} & 0.000 & -0.458 & 0.699 & 0.592 \\
 & 0.10 & \textbf{1.347} & 0.000 & 0.981 & 1.053 & 1.312 &  & 0.10 & \textbf{0.715} & 0.000 & 0.000 & 0.699 & 0.592 \\
 & 0.20 & \textbf{1.069} & 0.000 & 0.981 & 1.053 & 1.055 &  & 0.20 & \textbf{0.723} & 0.000 & 0.000 & 0.475 & 0.642 \\
 & 0.30 & 1.051 & 0.000 & 0.000 & 1.053 & \textbf{1.065} &  & 0.30 & \textbf{0.693} & 0.000 & 0.000 & 0.000 & 0.551 \\
\midrule
\multirow{4}{*}{Bikesharing} & 0.05 & \textbf{2.180} & 0.000 & 1.290 & 0.000 & 0.704 & \multirow{4}{*}{SML2010} & 0.05 & \textbf{1.867} & 0.000 & 0.000 & 1.180 & 1.063 \\
 & 0.10 & \textbf{1.557} & 0.000 & 1.290 & 0.000 & 0.704 &  & 0.10 & \textbf{1.384} & 0.000 & 0.000 & 1.180 & 1.014 \\
 & 0.20 & \textbf{1.048} & 0.000 & 0.000 & 0.000 & 0.704 &  & 0.20 & \textbf{1.045} & 0.000 & 0.000 & 0.000 & 0.639 \\
 & 0.30 & \textbf{0.757} & 0.000 & 0.000 & 0.000 & 0.694 &  & 0.30 & \textbf{0.728} & 0.000 & 0.000 & 0.000 & 0.000 \\
\midrule
\multirow{4}{*}{California} & 0.05 & \textbf{1.645} & 0.000 & 0.000 & 0.000 & 0.747 & \multirow{4}{*}{Supercond} & 0.05 & \textbf{1.744} & 0.000 & 0.839 & 0.773 & 0.665 \\
 & 0.10 & \textbf{1.292} & 0.000 & 0.000 & 0.000 & 0.747 &  & 0.10 & \textbf{1.278} & 0.000 & 0.839 & 0.773 & 0.665 \\
 & 0.20 & \textbf{0.904} & 0.000 & 0.000 & 0.000 & 0.444 &  & 0.20 & \textbf{0.972} & 0.000 & 0.839 & 0.773 & 0.665 \\
 & 0.30 & \textbf{0.635} & 0.000 & 0.000 & 0.000 & 0.251 &  & 0.30 & \textbf{0.798} & 0.000 & 0.000 & 0.000 & 0.665 \\
\midrule
\multirow{4}{*}{Concrete} & 0.05 & \textbf{1.278} & 1.174 & 0.000 & 1.220 & 1.213 & \multirow{4}{*}{Support2} & 0.05 & \textbf{0.145} & 0.000 & -0.181 & 0.000 & -0.163 \\
 & 0.10 & 0.593 & 0.996 & 0.000 & 0.993 & \textbf{1.213} &  & 0.10 & \textbf{0.055} & 0.000 & -0.181 & 0.000 & -0.199 \\
 & 0.20 & \textbf{0.478} & 0.000 & 0.000 & 0.000 & 0.186 &  & 0.20 & \textbf{0.036} & 0.000 & -0.181 & 0.000 & -0.110 \\
 & 0.30 & \textbf{0.382} & 0.000 & 0.000 & 0.000 & 0.000 &  & 0.30 & \textbf{0.038} & 0.000 & -0.181 & 0.000 & -0.110 \\
\midrule
\multirow{4}{*}{Elevators} & 0.05 & 0.318 & 0.000 & -0.268 & -0.100 & \textbf{0.447} & \multirow{4}{*}{Synchronous} & 0.05 & 1.375 & \textbf{1.525} & 0.000 & 1.204 & 0.965 \\
 & 0.10 & 0.318 & 0.000 & -0.240 & -0.100 & \textbf{0.447} &  & 0.10 & 1.375 & \textbf{1.525} & 0.000 & 1.204 & 0.965 \\
 & 0.20 & \textbf{0.318} & 0.000 & 0.000 & -0.100 & 0.258 &  & 0.20 & \textbf{1.157} & 0.000 & 0.000 & 0.000 & 0.673 \\
 & 0.30 & \textbf{0.318} & 0.000 & 0.000 & -0.100 & 0.263 &  & 0.30 & \textbf{0.824} & 0.000 & 0.000 & 0.000 & 0.789 \\
\midrule
\multirow{4}{*}{Energy} & 0.05 & \textbf{1.783} & 0.000 & 1.088 & 1.250 & 0.844 & \multirow{4}{*}{Thermography} & 0.05 & \textbf{0.946} & 0.554 & 0.000 & 0.000 & 0.451 \\
 & 0.10 & \textbf{1.544} & 0.000 & 1.088 & 1.250 & 0.844 &  & 0.10 & \textbf{0.700} & 0.636 & 0.000 & 0.000 & 0.406 \\
 & 0.20 & \textbf{1.138} & 0.000 & 1.056 & 0.000 & 0.844 &  & 0.20 & \textbf{0.548} & 0.349 & 0.000 & 0.000 & 0.000 \\
 & 0.30 & \textbf{0.909} & 0.000 & 0.000 & 0.000 & 0.844 &  & 0.30 & \textbf{0.319} & 0.289 & 0.000 & 0.000 & 0.000 \\
\midrule
\multirow{4}{*}{ForestFires} & 0.05 & 0.000 & 0.131 & 0.000 & 0.000 & \textbf{0.579} & \multirow{4}{*}{Toxicity} & 0.05 & \textbf{0.897} & 0.465 & 0.000 & 0.000 & 0.617 \\
 & 0.10 & 0.000 & 0.131 & 0.000 & 0.000 & \textbf{0.579} &  & 0.10 & \textbf{0.842} & 0.621 & 0.000 & 0.000 & 0.617 \\
 & 0.20 & 0.000 & 0.000 & 0.000 & 0.000 & 0.000 &  & 0.20 & \textbf{0.498} & 0.000 & 0.000 & 0.000 & 0.000 \\
 & 0.30 & 0.000 & 0.000 & 0.000 & 0.000 & 0.000 &  & 0.30 & \textbf{0.238} & 0.000 & 0.000 & 0.000 & 0.000 \\
\midrule
\multirow{4}{*}{House} & 0.05 & \textbf{1.186} & 0.000 & 0.000 & 0.000 & 0.447 & \multirow{4}{*}{Treasury} & 0.05 & 1.605 & 1.282 & 0.000 & \textbf{1.923} & 0.817 \\
 & 0.10 & \textbf{0.943} & 0.000 & 0.000 & 0.000 & 0.447 &  & 0.10 & \textbf{1.458} & 0.000 & 0.000 & 0.000 & 0.817 \\
 & 0.20 & \textbf{0.710} & 0.000 & 0.000 & 0.000 & 0.369 &  & 0.20 & \textbf{1.258} & 0.000 & 0.000 & 0.000 & 0.817 \\
 & 0.30 & \textbf{0.549} & 0.000 & 0.000 & 0.000 & 0.000 &  & 0.30 & \textbf{0.988} & 0.000 & 0.000 & 0.000 & 0.817 \\
\midrule
\multirow{4}{*}{NavalProp} & 0.05 & \textbf{1.551} & 0.000 & 0.000 & 0.120 & 0.003 & \multirow{4}{*}{Wizmir} & 0.05 & 1.516 & \textbf{1.801} & 1.629 & 1.072 & 1.016 \\
 & 0.10 & \textbf{1.344} & 0.000 & 0.000 & 0.000 & 0.003 &  & 0.10 & 1.565 & 1.226 & \textbf{1.629} & 1.072 & 1.016 \\
 & 0.20 & \textbf{1.128} & 0.000 & 0.000 & 0.000 & 0.003 &  & 0.20 & \textbf{1.151} & 0.000 & 0.000 & 1.072 & 1.016 \\
 & 0.30 & \textbf{0.871} & 0.000 & 0.000 & 0.000 & 0.003 &  & 0.30 & 0.841 & 0.000 & 0.000 & 0.000 & \textbf{0.920} \\
\midrule
\multirow{4}{*}{Puma32h} & 0.05 & \textbf{2.365} & 0.000 & 0.574 & 1.913 & 0.439 &  &  &  &  &  &  &  \\
 & 0.10 & \textbf{1.783} & 0.000 & 0.574 & 0.000 & 0.439 &  &  &  &  &  &  &  \\
 & 0.20 & \textbf{1.132} & 0.000 & 0.568 & 0.000 & 0.439 &  &  &  &  &  &  &  \\
 & 0.30 & \textbf{0.760} & 0.000 & 0.000 & 0.000 & 0.439 &  &  &  &  &  &  &  \\
\midrule
\#Best & 0.05 & \textbf{14} & 2 & 0 & 1 & 2 & Avg.\ Rank & 0.05 & \textbf{1.42} & 3.53 & 4.00 & 3.05 & 3.00 \\
 & 0.10 & \textbf{14} & 1 & 1 & 0 & 3 &  & 0.10 & \textbf{1.47} & 3.50 & 3.74 & 3.50 & 2.79 \\
 & 0.20 & \textbf{18} & 0 & 0 & 0 & 0 &  & 0.20 & \textbf{1.11} & 3.92 & 3.68 & 3.68 & 2.61 \\
 & 0.30 & \textbf{16} & 0 & 0 & 0 & 2 &  & 0.30 & \textbf{1.26} & 3.66 & 3.89 & 3.71 & 2.47 \\
\textit{Total} & -- & \textbf{62} & 3 & 1 & 1 & 7 & \textit{Overall} & -- & \textbf{1.32} & 3.65 & 3.83 & 3.49 & 2.72 \\
\bottomrule
\end{tabular}}
\end{table}
\subsection{Subgroup Improvement with support thresholds given as ratios} \label{appendix:subgroup_ratio_improvement}
\begin{figure}[h!]
    \centering
    \includegraphics[width=0.85\linewidth]{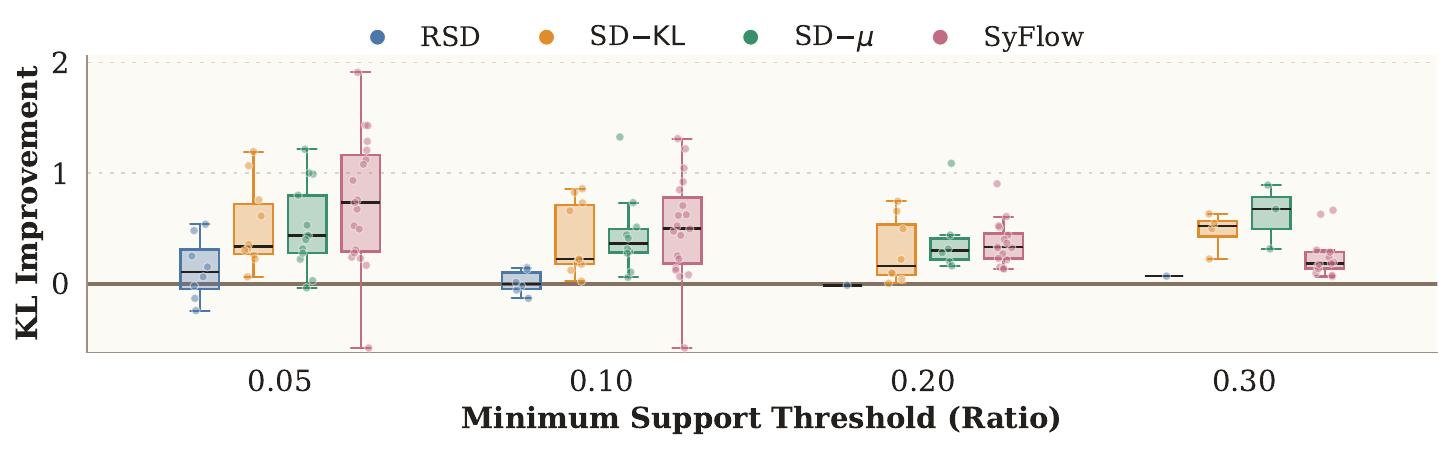}
    \caption{\camera{Held-out KL-divergence difference between AtomSD's refined subgroup and each competitor's original subgroup, across 19 datasets and four support ratios $\alpha'$. Positive values indicate that AtomSD's refinement improves the given subgroups.}}
    \label{fig:exp2_ratio}
\end{figure}

Figure~\ref{fig:exp2_ratio} reports the support-constrained subgroup improvement experiment under support thresholds expressed as ratios, complementing the absolute-count thresholds in the main text. For each dataset and each competitor, we apply our method as a post-processing step to the competitor's returned subgroup and report the difference in held-out KL divergence; positive values indicate that our method yields a more deviating subgroup \camera{within the same atoms}. Similarly, as Appendix~\ref{appendix:subsec:instability_kl} suggests that subgroups with less than 50 training instances will lead to estimated KL-divergence with prohibitively large variance, we skip subgroups for post-processing \camera{whose support (number of training instances) is smaller than $50$}. 

The pattern is consistent with the main-text results: the distribution of KL differences is \camera{predominantly} positive across all four competitors and all four support ratios\camera{. Medians are above zero for SyFlow, SD-KL and SD-$\mu$ at every ratio; for RSD, which returns a subgroup satisfying the constraint on only a few datasets beyond $\alpha' = 0.10$ (one or two per box), the median is around zero.} This reinforces the interpretation advanced in the main text: 1) the structural condition identified by our theorem (union of whole atoms plus at most one fractional atom) is empirically necessary, and 2) existing methods routinely return subgroups violating it, leaving KL divergence on the table that our post-processing step reliably recovers.

The magnitude of improvement varies systematically with the competitor. Gains over SyFlow are largest at small $\camera{\alpha'}$ and narrow as $\camera{\alpha'}$ grows, while gains over RSD are \camera{small or absent}. RSD often returns no subgroup at moderate-to-large ratios (see Table~\ref{table:results_exp3_ratio}), so there is less structural slack to exploit when it does. The consistency of the effect across competitors is the key observation: the improvement is not an artifact of any particular baseline but a systematic consequence of the structural condition being violated in practice.

\subsection{CD-diagrams and significance testing} \label{appendix:cd-diagram}
\begin{figure}
    \centering
    \includegraphics[width=0.48\linewidth]{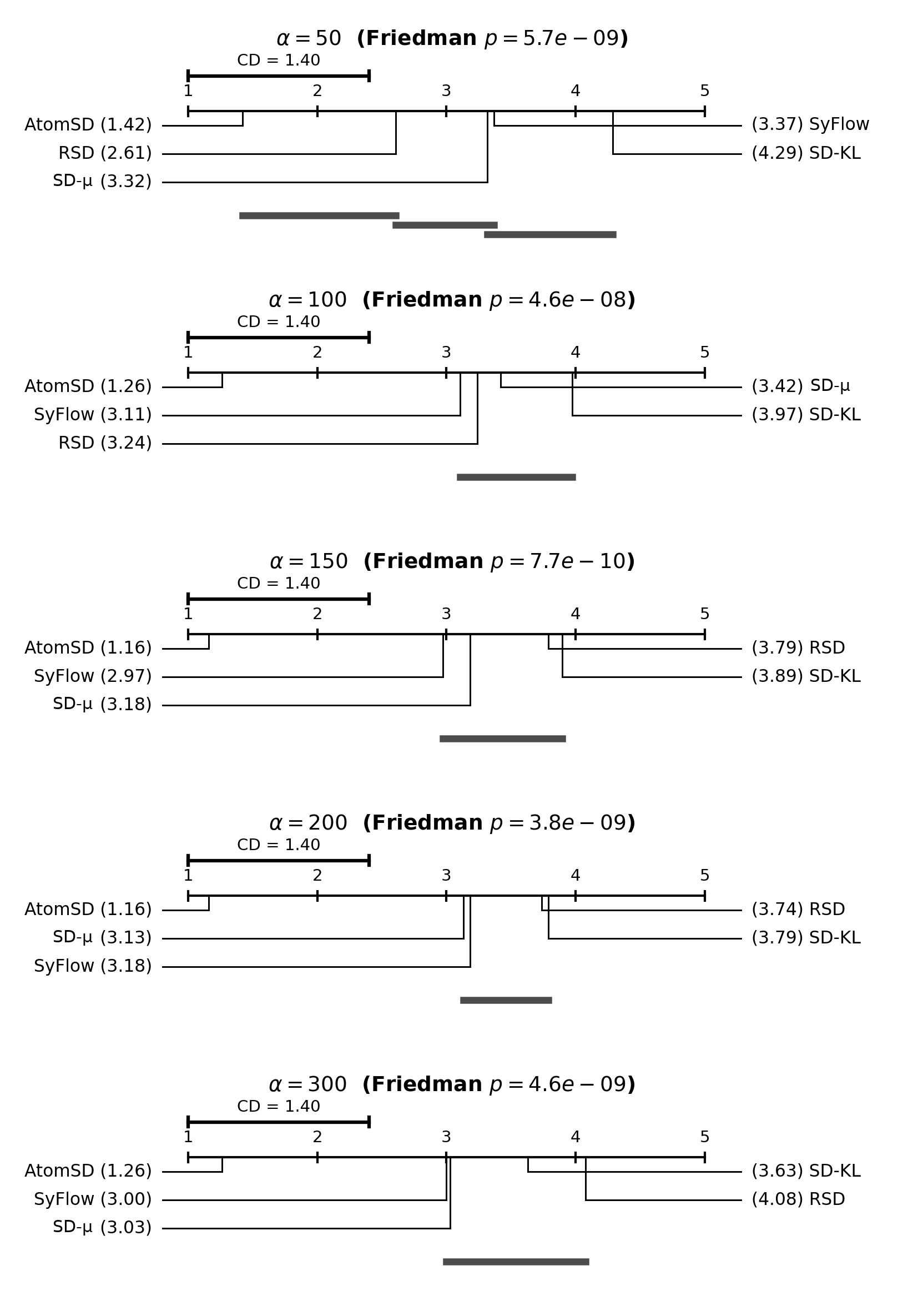}
    \includegraphics[width=0.48\linewidth]{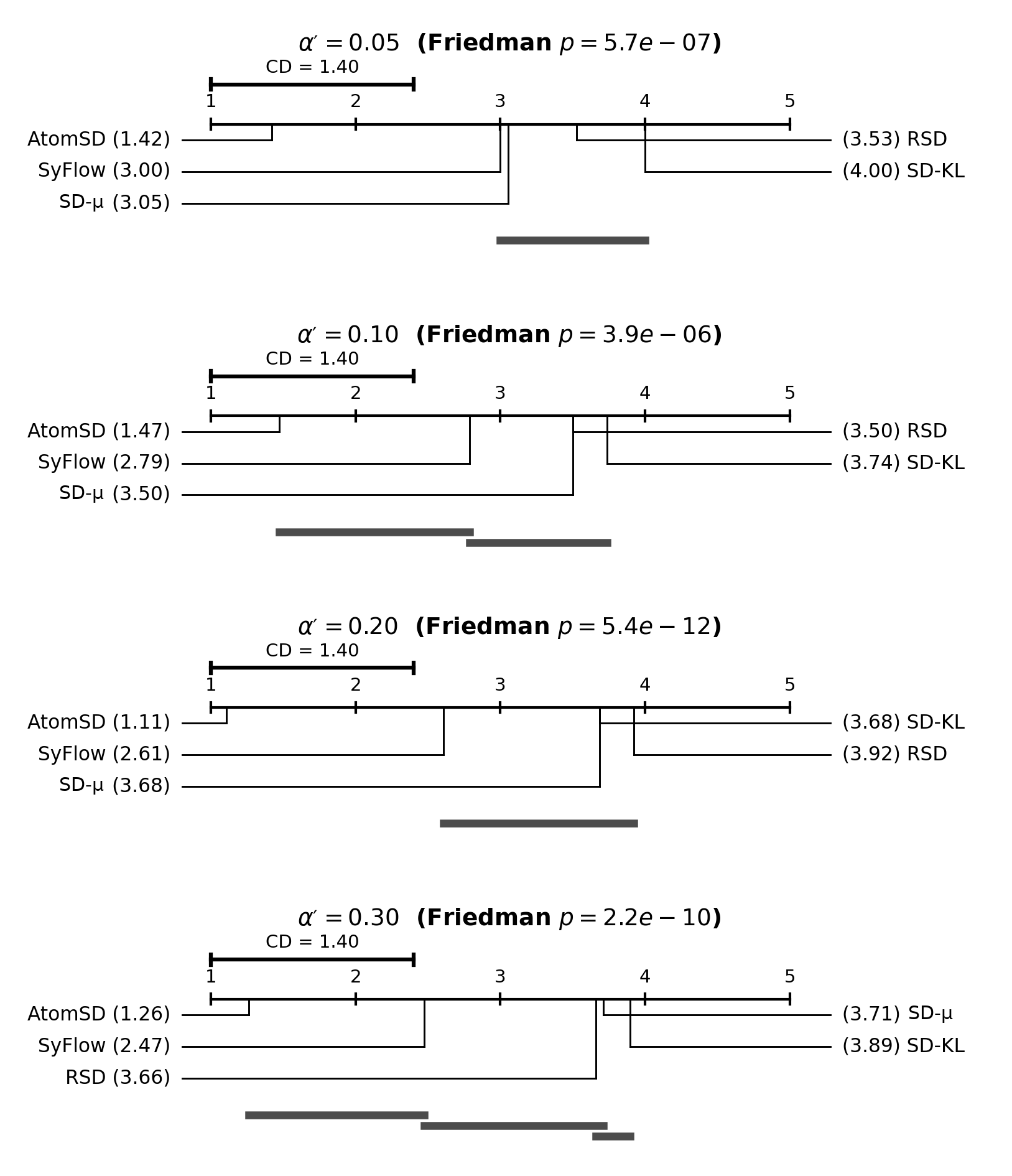}
       \caption{Critical-difference diagrams for constrained subgroup discovery across all support thresholds. \textbf{Left column:} absolute thresholds $\alpha \in \{50, 100, 150, 200, 300\}$ (Table~\ref{tab:exp3}). \textbf{Right column:} ratio thresholds $\camera{\alpha'} \in \{0.05, 0.10, 0.20, 0.30\}$ (Table~\ref{table:results_exp3_ratio}). Each diagram reports the Friedman test $p$-value \camera{(Iman--Davenport $F$ statistic)} and the Nemenyi critical difference (CD~$=1.40$ at significance level $0.05$). Methods connected by a horizontal bar are not significantly different; methods separated by more than the CD are significantly different.}
    \label{fig:cddiagram}
\end{figure}

Figure~\ref{fig:cddiagram} shows critical-difference diagrams~\citep{demvsar2006statistical} corresponding to the constrained subgroup discovery results in Tables~\ref{tab:exp3} and~\ref{table:results_exp3_ratio}. At every support threshold in both regimes, the Friedman test \camera{(Iman--Davenport $F$ statistic)} rejects the null hypothesis of equal performance at $p < 10^{-5}$, \camera{and AtomSD is ranked first in every setting with a mean rank of 1.11--1.47. Its separation from the runner-up exceeds the critical difference in six of the nine settings; in the remaining three ($\alpha = 50$, $\alpha' = 0.10$, $\alpha' = 0.30$) the gap (\camera{1.19}--1.32) falls slightly short of the critical difference of 1.40.} The relative ordering of the remaining methods is less stable across thresholds (e.g., RSD is the second-best at $\alpha = 50$ but drops substantially at larger $\alpha$), reflecting the baselines' sensitivity to the implicit support--deviation trade-off embedded in their respective quality measures. AtomSD's ranking, by contrast, is consistent across both absolute and ratio regimes, which we view as further evidence that the theory-directed atom-selection approach is not tuned to a particular support scale.

\subsection{Empirically validating Theorem~\ref{thm:optimality}: searching over the leaves/atomic subgroups for unconstrained subgroup discovery}
We empirically validate Theorem~\ref{thm:optimality} as a sanity check: since Theorem~\ref{thm:optimality} suggests that at the population level, the atomic subgroup with the largest KL-divergence is the most deviating subgroup, we empirically examine the KL-divergence of the atomic subgroups (i.e., the leaves) learned by CDTree. As our experiments in Appendix~\ref{appendix:subsec:instability_kl} suggests that subgroups with less than 50 training instances will lead to estimated KL-divergence with prohibitively large variance, we exclude any atomic subgroups that has less than 50 training instances in this comparison. 
Accordingly, all subgroups with less than 50 training instances from competitor algorithms are also removed. 

Table~\ref{tab:results_exp1} shows that our method delivers the strongest overall performance on unconstrained subgroup discovery. \camera{(Here AtomSD returns a single leaf, so its values can be below the constrained ones in Table~\ref{tab:exp3}, where GAP may return unions of leaves.)} It achieves the highest mean KL divergence on 10 of 19 datasets and obtains the best average rank of \camera{1.789}, ahead of all competitors. The margins are often substantial as well and it scales gracefully from low-dimensional problems (e.g., \camera{\textit{Energy}}, 4 features) to high-dimensional ones (e.g., \camera{\textit{Supercond}}, 81 features). RSD is the most consistent runner-up, winning on five datasets, but collapses on larger or harder instances such as \camera{\textit{California}, \textit{Elevators}, and \textit{House}}, where it \camera{returns a degenerate subgroup (KL $\le 0$) or none at all}. \camera{SD-KL and SD-$\mu$} are competitive only on a handful of low-dimensional datasets and produce degenerate KL values elsewhere, while SyFlow's performance is uneven and rarely leading. Taken together, the results indicate that our approach finds more deviating subgroups than existing baselines and, crucially, does so reliably across a wide range of dataset sizes and dimensionalities.
\begin{table}[ht]
\centering
\caption{Results for unconstrained subgroup discovery. Each cell reports the mean maximum KL divergence over three held-out folds (standard deviation in parentheses), measured between the discovered subgroup and the marginal distribution of the full held-out split. Per-dataset best means are bold. \#Best counts datasets on which each algorithm attains the highest mean; Avg. Rank is the mean rank across datasets (lower is better).}
\label{tab:results_exp1}
\resizebox{\textwidth}{!}{
\begin{tabular}{lrrrrrrr}
\toprule
Dataset & \# rows & dim. & \textbf{AtomSD} & RSD & \camera{SD-KL} & SD-$\mu$ & SyFlow \\    
\midrule
Ailerons & 13750 & 40 & \textbf{1.446} ($\pm$0.442) & 1.363 ($\pm$0.168) & 0.981 ($\pm$0.000) & 1.053 ($\pm$0.005) & 1.312 ($\pm$0.390) \\
Bikesharing & 17379 & 17 & \textbf{3.056} ($\pm$0.104) & 0.891 ($\pm$0.300) & 1.290 ($\pm$0.000) & 2.889 ($\pm$0.229) & 1.246 ($\pm$0.386) \\
California & 20640 & 8 & \textbf{1.962} ($\pm$0.209) & -0.694 ($\pm$0.000) & 0.000 ($\pm$0.000) & 1.316 ($\pm$0.108) & 0.747 ($\pm$0.072) \\
Concrete & 1030 & 8 & 1.147 ($\pm$0.163) & 1.174 ($\pm$0.201) & 0.000 ($\pm$0.000) & \textbf{1.220} ($\pm$0.224) & 1.213 ($\pm$0.190) \\
Elevators & 16599 & 18 & \textbf{1.509} ($\pm$0.096) & 0.000 ($\pm$0.000) & -0.268 ($\pm$0.244) & -0.100 ($\pm$0.063) & 0.447 ($\pm$0.209) \\
Energy & 9568 & 4 & 2.105 ($\pm$0.083) & \textbf{2.229} ($\pm$0.138) & 1.088 ($\pm$0.035) & 1.250 ($\pm$0.020) & 0.844 ($\pm$0.067) \\
ForestFires & 517 & 12 & 0.000 ($\pm$0.000) & 0.131 ($\pm$0.000) & 0.000 ($\pm$0.000) & 0.000 ($\pm$0.000) & \textbf{0.579} ($\pm$0.000) \\
House & 22784 & 16 & \textbf{1.697} ($\pm$0.291) & 0.000 ($\pm$0.000) & 0.000 ($\pm$0.000) & 1.501 ($\pm$0.107) & 0.447 ($\pm$0.112) \\
NavalProp & 11934 & 16 & \textbf{1.687} ($\pm$0.093) & 1.242 ($\pm$0.283) & 0.878 ($\pm$0.542) & 0.449 ($\pm$0.582) & 0.952 ($\pm$0.358) \\
Puma32h & 8192 & 32 & 2.769 ($\pm$0.230) & \textbf{3.088} ($\pm$0.292) & 0.574 ($\pm$0.008) & 1.913 ($\pm$0.068) & 0.439 ($\pm$0.009) \\
SkillCraft & 3338 & 18 & 0.387 ($\pm$0.362) & \textbf{1.035} ($\pm$0.000) & -0.458 ($\pm$0.647) & 0.699 ($\pm$0.042) & 0.592 ($\pm$0.170) \\
SML2010 & 2764 & 20 & \textbf{2.152} ($\pm$0.314) & 2.029 ($\pm$0.000) & 0.000 ($\pm$0.000) & 1.180 ($\pm$0.186) & 1.253 ($\pm$0.432) \\
Supercond & 21263 & 81 & \textbf{3.094} ($\pm$0.183) & 0.855 ($\pm$0.089) & 0.839 ($\pm$0.017) & 0.773 ($\pm$0.019) & 0.665 ($\pm$0.028) \\
Support2 & 9105 & 26 & \textbf{0.729} ($\pm$0.077) & 0.119 ($\pm$0.513) & -0.181 ($\pm$0.005) & -0.175 ($\pm$0.137) & -0.163 ($\pm$0.070) \\
Synchronous & 557 & 4 & 1.205 ($\pm$0.000) & \textbf{1.525} ($\pm$0.000) & 0.000 ($\pm$0.000) & 1.204 ($\pm$0.297) & 0.965 ($\pm$0.453) \\
Thermography & 1003 & 25 & \textbf{0.911} ($\pm$0.350) & 0.554 ($\pm$0.185) & 0.000 ($\pm$0.000) & 0.000 ($\pm$0.000) & 0.451 ($\pm$0.219) \\
Toxicity & 908 & 6 & 0.576 ($\pm$0.179) & 0.465 ($\pm$0.201) & 0.000 ($\pm$0.000) & 0.000 ($\pm$0.000) & \textbf{0.617} ($\pm$0.372) \\
Treasury & 1049 & 15 & 1.454 ($\pm$0.310) & 1.282 ($\pm$0.000) & 0.000 ($\pm$0.000) & \textbf{1.923} ($\pm$0.090) & 0.817 ($\pm$0.121) \\
Wizmir & 1461 & 9 & 1.656 ($\pm$0.122) & \textbf{1.801} ($\pm$0.204) & 1.629 ($\pm$0.000) & 1.072 ($\pm$0.088) & 1.016 ($\pm$0.171) \\
\midrule
\#Best & & & \textbf{10} & 5 & 0 & 2 & 2 \\
Avg. Rank & & & \textbf{1.789} & 2.395 & 4.342 & 3.211 & 3.263 \\ 
\bottomrule
\end{tabular}
}
\end{table}

\section{Additional Algorithmic Details} \label{app:sec:alg}
We provide further details of the proposed optimization algorithms. 
\paragraph{SearchFull.}
Algorithm~\ref{alg:search full} specifies the \textsc{SearchFull} subroutine, which refines the \camera{index set $\mathcal{I}$} of fully-included atoms while keeping the partial component $p$ fixed. The procedure is a standard local search over the power set of candidate atoms: at each iteration it considers every single-atom toggle --- adding an atomic subgroup $\camera{S_i}$ currently absent from $\camera{\mathcal{I}}$, or removing one currently present --- and commits to the best improving move, subject to the minimum-support constraint $\alpha$. Candidates whose support overlaps with the current partial component $p$ are excluded (line~4) to avoid double-counting instances across the full and partial parts of the subgroup. The loop terminates when no single-atom toggle improves the objective, yielding a locally optimal selection of whole atoms given the current $p$.

The move set is deliberately restricted to single-atom flips. Larger move sets (e.g., pairwise swaps) would enlarge the neighbourhood at the cost of per-iteration complexity, and we found empirically that the single-flip neighbourhood already yields competitive solutions when alternated with the partial-refinement step \textsc{SearchPartial}. By Theorem~\ref{thm:support_existence}, the globally optimal subgroup has the form $(\bigcup_{i \in \camera{\mathcal{I}}} S_i) \cup S_k'$ for some index set $\camera{\mathcal{I}}$ and at most one partial atom $S_k'$; \textsc{SearchFull} targets the $\camera{\mathcal{I}}$ component of this decomposition, and \textsc{SearchPartial} targets the $S_k'$ component. Alternating the two (Algorithm~\ref{alg:expansion-union}) therefore operates directly in the solution class guaranteed by the theorem.
\begin{algorithm}[ht]
\caption{SearchFull: Search for Unions of Full Atomic Subgroups}
\label{alg:search full}
\begin{algorithmic}[1]\cameracolor
\Require Current index set $\mathcal{I} \subseteq [K]$ of whole atoms and partial subgroup $p$; candidate atoms $\mathcal{S}$, training data, minimum support $\alpha$, \Call{Score}{} and \Call{supp}{} as in Algorithm~\ref{alg:expansion-union}.
\Ensure The index set $\mathcal{I}$ after a local search over single-atom toggles, with $p$ unchanged.
\Statex
\Function{SearchFull}{$\mathcal{I}, p$}
    \Repeat
        \State $\mathcal{I}_{\mathrm{best}} \gets \mathcal{I}$
        \ForAll{$i \in [K]$ such that $p \cap S_i = \varnothing$} \Comment{Consider toggling atom $i$}
            \State $\mathcal{I}' \gets \mathcal{I} \setminus \{i\}$ if $i \in \mathcal{I}$, otherwise $\mathcal{I} \cup \{i\}$
            \If{$\Call{supp}{\mathcal{I}', p} < \alpha$} \textbf{continue} \Comment{Support constraint violated}
            \EndIf
            \If{$\Call{Score}{\mathcal{I}', p} > \Call{Score}{\mathcal{I}_{\mathrm{best}}, p}$} $\mathcal{I}_{\mathrm{best}} \gets \mathcal{I}'$
            \EndIf
        \EndFor
        \State $\mathit{improved} \gets \Call{Score}{\mathcal{I}_{\mathrm{best}}, p} > \Call{Score}{\mathcal{I}, p}$
        \If{$\mathit{improved}$} $\mathcal{I} \gets \mathcal{I}_{\mathrm{best}}$ \Comment{Commit the best improving toggle}
        \EndIf
    \Until{$\lnot\, \mathit{improved}$}
    \State \Return $(\mathcal{I}, p)$
\EndFunction
\end{algorithmic}
\end{algorithm}

\paragraph{SearchPartial.}
Algorithm~\ref{alg:search partial} specifies the \textsc{SearchPartial} subroutine, which refines the partial component $p$ while keeping the \camera{index set $\mathcal{I}$} fixed in structure. \camera{For each candidate atom $S_i$, the procedure uses $S_i$ as the partial atom: $i$ is removed from the index set if present (line~7) and the bounding box of $S_i$ is shrunk into a partial region $p' \subseteq S_i$. The best resulting $(\mathcal{I}', p')$ across all candidate atoms is returned.}

\qh{The core of the procedure is a best-improvement shrinkage of $\camera{S_i}$'s bounding box (lines~\camera{9--19}). A \emph{cut} tightens exactly one bound of one feature: it raises the lower bound, or lowers the upper bound, of that feature to a value observed within the current box. Features fixed by equality constraints in $\camera{S_i}$ are excluded. Starting from the full bounds $[\boldsymbol{\ell}, \boldsymbol{u}]$ of $\camera{S_i}$, each iteration scores every admissible cut and commits the one that most improves $\mathrm{score}$ on the union with \camera{the atoms in $\mathcal{I}'$}, subject to the support constraint. \camera{If no cut is ever applied, $S_i$ is kept as a whole atom (line~21).} The loop terminates when no cut improves the objective.}

Two design choices deserve comment. First, restricting cuts to observed values rather than arbitrary thresholds keeps the candidate pool finite and ensures every partial region corresponds to a valid rule on the training data. 
\qh{Second, cuts are chosen by best improvement rather than first improvement: all candidate cuts across all features are scored before the box changes, so the result does not depend on the order in which features are visited. The search stops at a local optimum with respect to such cuts and does not guarantee a globally optimal partial region.}
The output $\camera{p'} = [\boldsymbol{\ell}, \boldsymbol{u}]$ is a rectangular partial region within $\camera{S_i}$, exactly the form of $S_k'$ in the decomposition guaranteed by Theorem~\ref{thm:support_existence}.
\begin{algorithm}[ht]
\caption{SearchPartial: Search for the Optional Partial Atomic Subgroup}
\label{alg:search partial}
\begin{algorithmic}[1]\cameracolor
\Require Current index set $\mathcal{I} \subseteq [K]$ of whole atoms and partial subgroup $p$; candidate atoms $\mathcal{S}$, training data, minimum support $\alpha$, \Call{Score}{} and \Call{supp}{} as in Algorithm~\ref{alg:expansion-union}; $\Call{Box}{S_i}$: the bounding box $[\boldsymbol{\ell}, \boldsymbol{u}]$ of $S_i$ over its non-equality-constrained features; \Call{Convert}{$\boldsymbol{\ell}, \boldsymbol{u}$}: the rule (sub-region of $S_i$) with these bounds.
\Ensure The best-scoring pair among the current $(\mathcal{I}, p)$, the pruned pair $(\mathcal{I}, \varnothing)$, and one candidate per atom $i$: either $(\mathcal{I} \setminus \{i\},\, p')$ with a shrunk $p' \subsetneq S_i$, or $(\mathcal{I} \cup \{i\}, \varnothing)$ if no cut is applied; all subject to the minimum support $\alpha$.
\Statex
\Function{SearchPartial}{$\mathcal{I}, p$}
    \State $(\mathcal{I}_{\mathrm{best}}, p_{\mathrm{best}}) \gets (\mathcal{I}, p)$
    \If{$\Call{supp}{\mathcal{I}, \varnothing} \geq \alpha$ \textbf{and} $\Call{Score}{\mathcal{I}, \varnothing} > \Call{Score}{\mathcal{I}, p}$}
        \State $(\mathcal{I}_{\mathrm{best}}, p_{\mathrm{best}}) \gets (\mathcal{I}, \varnothing)$ \Comment{Dropping $p$ is better}
    \EndIf
    \ForAll{$i \in [K]$} \Comment{Try $S_i$ as the partial atom}
        \State $\mathcal{I}' \gets \mathcal{I} \setminus \{i\}$ \Comment{Unchanged if $i \notin \mathcal{I}$}
        \State $[\boldsymbol{\ell}, \boldsymbol{u}] \gets \Call{Box}{S_i}$ \Comment{Start from the full extent of $S_i$}
        \Repeat \Comment{Best-improvement shrinkage: each iteration tightens one bound of one feature}
            \State $p' \gets \Call{Convert}{\boldsymbol{\ell}, \boldsymbol{u}}$; \quad $c_{\mathrm{best}} \gets \varnothing$; \quad $s_{\mathrm{best}} \gets \Call{Score}{\mathcal{I}', p'}$
            \ForAll{cuts $c \in \{\ell_f \gets v,\ u_f \gets v\}$, $v$ a value of feature $f$ observed in $p'$}
                \State $p_c \gets \Call{Convert}{[\boldsymbol{\ell}, \boldsymbol{u}] \text{ with cut } c \text{ applied}}$
                \If{$\Call{supp}{\mathcal{I}', p_c} \geq \alpha$ \textbf{and} $\Call{Score}{\mathcal{I}', p_c} > s_{\mathrm{best}}$}
                    \State $(c_{\mathrm{best}}, s_{\mathrm{best}}) \gets (c, \Call{Score}{\mathcal{I}', p_c})$
                \EndIf
            \EndFor
            \If{$c_{\mathrm{best}} \neq \varnothing$} apply $c_{\mathrm{best}}$ to $[\boldsymbol{\ell}, \boldsymbol{u}]$
            \EndIf
        \Until{$c_{\mathrm{best}} = \varnothing$}
        \State $p' \gets \Call{Convert}{\boldsymbol{\ell}, \boldsymbol{u}}$
        \If{$p' = S_i$} $(\mathcal{I}', p') \gets (\mathcal{I}' \cup \{i\}, \varnothing)$ \Comment{No cut applied: keep $S_i$ as a whole atom}
        \EndIf
        \If{$\Call{supp}{\mathcal{I}', p'} \geq \alpha$ \textbf{and} $\Call{Score}{\mathcal{I}', p'} > \Call{Score}{\mathcal{I}_{\mathrm{best}}, p_{\mathrm{best}}}$}
            \State $(\mathcal{I}_{\mathrm{best}}, p_{\mathrm{best}}) \gets (\mathcal{I}', p')$
        \EndIf
    \EndFor
    \State \Return $(\mathcal{I}_{\mathrm{best}}, p_{\mathrm{best}})$
\EndFunction
\end{algorithmic}
\end{algorithm}

\section{NP-hardness of finding the global-optimal union of atomic subgroups} \label{appendix:nphard}
\begin{restatable}[Worst-case NP-hardness]{theorem}{nphardthm}\label{thm:nphard}
Let $\{S_i\}_{i=1}^m$ partition $\mathcal{X}$. Given a threshold $\alpha \in (0,1)$, the problem
\begin{equation}
\label{eq:main-problem}
    \max_{I \subseteq [m]}\; \mathrm{KL}\!\left(P(Y \mid X \in \textstyle\bigcup_{i \in I} S_i) \,\Big\|\, P(Y)\right) \quad \text{subject to} \quad P_X\!\left(\textstyle\bigcup_{i \in I} S_i\right) \geq \alpha
\end{equation}
is NP-hard in general. This hardness holds even when $Y$ is continuous and each $P(Y \mid X \in S_i)$ is a uniform distribution on an interval disjoint from all others.
\end{restatable}
\begin{proof}
We prove the claim by a polynomial-time reduction from \textsc{Subset-Sum}. The key idea is to construct an instance where the KL divergence of any union of atoms depends only on its total weight, so that maximizing KL subject to a support constraint is equivalent to finding a subset of atoms with a prescribed total weight, which is precisely \textsc{Subset-Sum}.

\paragraph{Step 1: Recall the source problem.}
An instance of \textsc{Subset-Sum} consists of positive integers $a_1,\dots,a_m$ and a target $T \in \mathbb{N}$, and asks whether there exists $I \subseteq [m]$ such that $\sum_{i \in I} a_i = T$. This problem is NP-complete.

\paragraph{Step 2: Construct the reduction.}
Given a \textsc{Subset-Sum} instance $(a_1,\dots,a_m,T)$, let $A := \sum_{i=1}^m a_i$ and set $w_i := a_i/A$ and $\alpha := T/A$. \camera{Let $X$ be uniform on $[0,1]$ and} construct a partition $\{S_i\}_{i=1}^m$ of $[0,1]$ into consecutive intervals with $P_X(S_i) = w_i$, and assign conditional distributions $P(Y \mid X \in S_i) = U([i, i+1])$. These uniform distributions have pairwise disjoint supports by construction, which will be the key property enabling the KL computation below.

\paragraph{Step 3: Compute the KL divergence.}
Fix any $I \subseteq [m]$\camera{, write $A_I := \bigcup_{i \in I} S_i$,} and let $W_I := P_X(A_I) = \sum_{i \in I} w_i$. By the law of total probability, $P(Y \mid X \in A_I) = \sum_{i \in I} (w_i/W_I)\, U([i,i+1])$ and $P(Y) = \sum_{i=1}^m w_i\, U([i,i+1])$. Since the supports $[i,i+1]$ are pairwise disjoint, for $y \in [i,i+1]$ with $i \in I$ the densities simplify to $p_I(y) = w_i/W_I$ and $p(y) = w_i$, so the likelihood ratio $p_I(y)/p(y) = 1/W_I$ is constant across the entire support of $p_I$. This gives the clean expression
\[
\mathrm{KL}\!\left(P(Y \mid X \in A_I) \,\|\, P(Y)\right) = \log \frac{1}{W_I}.
\]
The KL divergence therefore depends on $I$ only through its total weight $W_I$ and, more importantly, not through which specific atoms are selected.

\paragraph{Step 4: Reduce to \textsc{Subset-Sum}.}
Since $w \mapsto \log(1/w)$ is strictly decreasing, the optimization problem~\eqref{eq:main-problem} reduces to finding $I$ with $W_I \geq \alpha$ that minimizes $W_I$. The decision version with threshold $\tau := \log(1/\alpha)$ therefore asks whether there exists $I$ with $W_I \geq \alpha$ and $\log(1/W_I) \geq \log(1/\alpha)$, i.e., $W_I \leq \alpha$. Together these two conditions force $W_I = \alpha$ exactly. Translating back via $w_i = a_i/A$ and $\alpha = T/A$, we have
\[
W_I = \alpha \quad \Longleftrightarrow \quad \sum_{i \in I} a_i = T,
\]
which is precisely the \textsc{Subset-Sum} decision problem. The reduction runs in polynomial time, so the decision version of~\eqref{eq:main-problem} is NP-hard, and hence so is the optimization problem itself. Since the construction uses only continuous $Y$ with uniform conditional distributions on pairwise disjoint intervals, the hardness holds already for this restricted class of instances.
\end{proof}
\clearpage
\section*{NeurIPS Paper Checklist}
\begin{enumerate}

\item {\bf Claims}
    \item[] Question: Do the main claims made in the abstract and introduction accurately reflect the paper's contributions and scope?
    \item[] Answer: \answerYes{}
    \item[] Justification: The abstract and introduction state three contributions: (i) a structural characterization of KL-optimal subgroups under support constraints (Theorem~\ref{thm:support_existence}, Corollary~\ref{cor:uniqueness}); (ii) a modular two-step recipe reducing the search to a finite combinatorial problem; and (iii) an instantiation (AtomSD) that empirically dominates existing methods across 19 UCI datasets and improves subgroups returned by competing methods as a post-processing step. Each claim is supported by the corresponding theorem in Section~\ref{sec:theory} or experiment in Section~\ref{sec:experiments}. The scope of the theoretical results is stated explicitly (partition-based generative model; distinct generating distributions for the necessity direction), and the introduction flags that real data deviates from this model to varying degrees.
 

\item {\bf Limitations}
    \item[] Question: Does the paper discuss the limitations of the work performed by the authors?
    \item[] Answer: \answerYes{}
    \item[] Justification: Section~\ref{sec:conclusion} contains a dedicated ``Limitations and future work'' paragraph that addresses (i) the partition-based generative model is an approximation of real data and formal robustness to distributional misspecification remains open, and (ii) the natural extension to subgroup-set discovery with diversity considerations. In addition, the experiments section explicitly discusses the finite-sample regime where the partition is learned rather than given (Section~\ref{sec:experiments}). 

\item {\bf Theory assumptions and proofs}
    \item[] Question: For each theoretical result, does the paper provide the full set of assumptions and a complete (and correct) proof?
    \item[] Answer: \answerYes{}
        \item[] Justification: All theoretical results (Theorem~\ref{thm:homogeneity}, Theorem~\ref{thm:optimality}, Theorem~\ref{thm:support_existence}, Corollary~\ref{cor:uniqueness}, and Theorem~\ref{thm:nphard}) are numbered, cross-referenced, and stated with their full set of assumptions. The partition-based generative model is formally introduced in Definition~\ref{def:partition_model}, and \camera{results requiring additional assumptions (a unique maximizing atom in Theorem~\ref{thm:optimality}; distinct generating distributions in Corollary~\ref{cor:uniqueness}) state them explicitly}. Complete proofs for all theorems are provided in Appendix~\ref{app:proofs}. 

    \item {\bf Experimental result reproducibility}
    \item[] Question: Does the paper fully disclose all the information needed to reproduce the main experimental results of the paper to the extent that it affects the main claims and/or conclusions of the paper (regardless of whether the code and data are provided or not)?
    \item[] Answer: \answerYes{}
        \item[] Justification: Section~\ref{sec:experiments} specifies the datasets (19 \camera{benchmark} datasets), competitor algorithms, evaluation protocol (3-fold cross-validation with held-out KL estimation), and dataset preprocessing. Full reproducibility details including hyperparameter choices, KL estimation procedure, and dataset summary are in Appendix~\ref{app:sec:setup}. Pseudocode for all algorithmic components (GAP, SearchFull, SearchPartial) is given in the main text (Algorithm~\ref{alg:expansion-union}) and Appendix~\ref{app:sec:alg} (Algorithms~\ref{alg:search full} and~\ref{alg:search partial}). Source code is provided at an anonymous URL.

\item {\bf Open access to data and code}
    \item[] Question: Does the paper provide open access to the data and code, with sufficient instructions to faithfully reproduce the main experimental results, as described in supplemental material?
        \item[] Answer: \answerYes{}
    \item[] Justification: Source code is available at \url{https://anonymous.4open.science/r/SubgroupDiscovery}. All datasets are publicly available\camera{, mostly from the UCI Machine Learning Repository~\citep{uci_link}}. 
    

\item {\bf Experimental setting/details}
    \item[] Question: Does the paper specify all the training and test details (e.g., data splits, hyperparameters, how they were chosen, type of optimizer) necessary to understand the results?
        \item[] Answer: \answerYes{}
    \item[] Justification: Data splits (3-fold cross-validation), hyperparameter settings (default values for all competitor methods; no hyperparameter tuning for AtomSD), evaluation metric (held-out KL divergence via histogram density estimation with Freedman--Diaconis binning and Bayesian smoothing), and support threshold ranges (absolute counts $\alpha \in \{50, 100, 150, 200, 300\}$ in the main text; ratios $\alpha' \in \{0.05, 0.10, 0.20, 0.30\}$ in Appendix~\ref{appendix:ratio_constrain_subgroup_discovery}) are all specified.


\item {\bf Experiment statistical significance}
    \item[] Question: Does the paper report error bars suitably and correctly defined or other appropriate information about the statistical significance of the experiments?
        \item[] Answer: \answerYes{}
    \item[] Justification: Table~\ref{tab:exp3} reports mean KL divergence across 3-fold cross-validation. A Friedman test is applied to test the null hypothesis of equal performance across methods, with $p < 10^{-5}$ at every support threshold; the corresponding Nemenyi critical-difference diagrams are provided in Appendix~\ref{appendix:cd-diagram}. Additionally, Appendix~\ref{appendix:subsec:instability_kl} reports a bootstrap stability analysis ($B=300$ replicates) characterizing the estimation variability of KL divergence for subgroups of varying sizes.

\item {\bf Experiments compute resources}
    \item[] Question: For each experiment, does the paper provide sufficient information on the computer resources (type of compute workers, memory, time of execution) needed to reproduce the experiments?
        \item[] Answer: \answerYes{}
    \item[] Justification: Appendix~\ref{app:sec:setup} states that all experiments were conducted on a cluster equipped with AMD EPYC 7662 64-Core Processors. The method is CPU-only and does not require GPU resources.

    
\item {\bf Code of ethics}
    \item[] Question: Does the research conducted in the paper conform, in every respect, with the NeurIPS Code of Ethics \url{https://neurips.cc/public/EthicsGuidelines}?
        \item[] Answer: \answerYes{}
    \item[] Justification: The research uses only publicly available benchmark datasets, does not involve human subjects, and does not release models or data with misuse risk.


\item {\bf Broader impacts}
    \item[] Question: Does the paper discuss both potential positive societal impacts and negative societal impacts of the work performed?
        \item[] Answer: \answerNA{}
    \item[] Justification: This is a foundational theoretical/methodological paper on subgroup discovery. The theoretical contribution (structural characterization of KL-optimal subgroups under support constraints) and algorithmic instantiation are domain-agnostic tools for exploratory data analysis. We do not anticipate direct positive or negative societal impact beyond the general benefits and risks of data-analysis tools, which are not specific to this work.

    
\item {\bf Safeguards}
    \item[] Question: Does the paper describe safeguards that have been put in place for responsible release of data or models that have a high risk for misuse (e.g., pre-trained language models, image generators, or scraped datasets)?
        \item[] Answer: \answerNA{}
    \item[] Justification: The paper does not release data or models that pose a high risk for misuse. The released code is an implementation of a domain-agnostic subgroup discovery method applied to publicly available benchmarks.


\item {\bf Licenses for existing assets}
    \item[] Question: Are the creators or original owners of assets (e.g., code, data, models), used in the paper, properly credited and are the license and terms of use explicitly mentioned and properly respected?
        \item[] Answer: \answerYes{}
    \item[] Justification: All datasets used are publicly available\camera{, mostly from the UCI Machine Learning Repository~\citep{uci_link},} and are cited accordingly. Competitor methods (SyFlow~\citep{xu2024learning}, RSD~\citep{proencca2022robust}, SD-$\mu$ and SD-KL via pysubgroup~\citep{lemmerich2018pysubgroup}, CDTree~\citep{yang2024conditional}) are used with their original implementations and cited at each point of use.


\item {\bf New assets}
    \item[] Question: Are new assets introduced in the paper well documented and is the documentation provided alongside the assets?
        \item[] Answer: \answerYes{}
    \item[] Justification: The AtomSD implementation is released at the anonymous repository \url{https://anonymous.4open.science/r/SubgroupDiscovery}, with documentation including usage instructions and reproduction scripts for the experiments reported in the paper.


\item {\bf Crowdsourcing and research with human subjects}
    \item[] Question: For crowdsourcing experiments and research with human subjects, does the paper include the full text of instructions given to participants and screenshots, if applicable, as well as details about compensation (if any)? 
        \item[] Answer: \answerNA{}
    \item[] Justification: The paper does not involve crowdsourcing or research with human subjects.


\item {\bf Institutional review board (IRB) approvals or equivalent for research with human subjects}
    \item[] Question: Does the paper describe potential risks incurred by study participants, whether such risks were disclosed to the subjects, and whether Institutional Review Board (IRB) approvals (or an equivalent approval/review based on the requirements of your country or institution) were obtained?
        \item[] Answer: \answerNA{}
    \item[] Justification: The paper does not involve research with human subjects.


\item {\bf Declaration of LLM usage}
    \item[] Question: Does the paper describe the usage of LLMs if it is an important, original, or non-standard component of the core methods in this research? Note that if the LLM is used only for writing, editing, or formatting purposes and does \emph{not} impact the core methodology, scientific rigor, or originality of the research, declaration is not required.
        \item[] Answer: \answerNA{}
    \item[] Justification: All theoretical results, algorithmic design, experimental protocols, and scientific interpretations were developed entirely by the authors. LLMs did not contribute to the core methodology, scientific rigor, or originality of the research.

\end{enumerate}

\end{document}